\documentclass[sigconf]{acmart}
\usepackage{bm}
\usepackage{amsmath}
\usepackage{xspace}
\usepackage{diagbox}
\usepackage{color}
\usepackage{makecell}

\AtBeginDocument{%
  \providecommand\BibTeX{{%
    \normalfont B\kern-0.5em{\scshape i\kern-0.25em b}\kern-0.8em\TeX}}}

\setcopyright{acmcopyright}
\copyrightyear{2020}
\acmYear{2020}
\acmDOI{10.1145/3394171.3413770}

\acmConference[MM '20] {Proceedings of the 28th ACM International Conference on Multimedia}{October 12--16, 2020}{Seattle, WA, USA.}
\acmBooktitle{Proceedings of the 28th ACM International Conference on Multimedia (MM '20), October 12--16, 2020, Seattle, WA, USA.}
\acmPrice{15.00}
\acmISBN{978-1-4503-7988-5/20/10}
\begin{document}
\fancyhead{}

\title{Anisotropic Stroke Control for Multiple Artists\\ Style Transfer}










\author{Xuanhong Chen$^{*}$, Xirui Yan$^{*}$, Naiyuan Liu$^{\dagger}$, Ting Qiu and Bingbing Ni$^{\boxtimes}$}
\thanks{$^\ast$Equal contribution.}
\thanks{$^\dagger$Contributed to the work while he was a research assistant at Shanghai Jiao Tong University.}
\thanks{$^\boxtimes$Corresponding author: Bingbing Ni.}
\affiliation{%
	\institution{Shanghai Jiao Tong University, Shanghai, China}
}
\email{{chen19910528,xiruiYan,776398420,nibingbing}@sjtu.edu.cn,liunaiyuan27@gmail.com}

\renewcommand{\thefootnote}{\fnsymbol{footnote}}

\renewcommand{\shortauthors}{Chen and Yan, et al.}

\begin{abstract}
Though significant progress has been made in artistic style transfer,
semantic information is usually difficult to be preserved in a fine-grained locally consistent manner by most existing methods,
especially when multiple artists styles are required to transfer within one single model.
To circumvent this issue, we propose a \emph{Stroke Control Multi-Artist Style Transfer} framework.
On the one hand, we design an \emph{Anisotropic Stroke Module (ASM)} which realizes the dynamic adjustment of style-stroke between the non-trivial and the trivial regions.
\emph{ASM} endows the network with the ability of adaptive semantic-consistency among various styles.
On the other hand, we present an novel \emph{Multi-Scale Projection Discriminator} to realize the texture-level conditional generation.
In contrast to the single-scale conditional discriminator, our discriminator is able to capture multi-scale texture clue to effectively distinguish a wide range of artistic styles.
Extensive experimental results well demonstrate the feasibility and effectiveness of our approach.
Our framework can transform a photograph into different artistic style oil painting via only ONE single model.
Furthermore, the results are with distinctive artistic style and retain the anisotropic semantic information.
The code is already available on github: https://github.com/neuralchen/ASMAGAN.
\end{abstract}



\maketitle

\section{Introduction}
\begin{figure}[t]
\begin{center}
\includegraphics[width=1\linewidth]{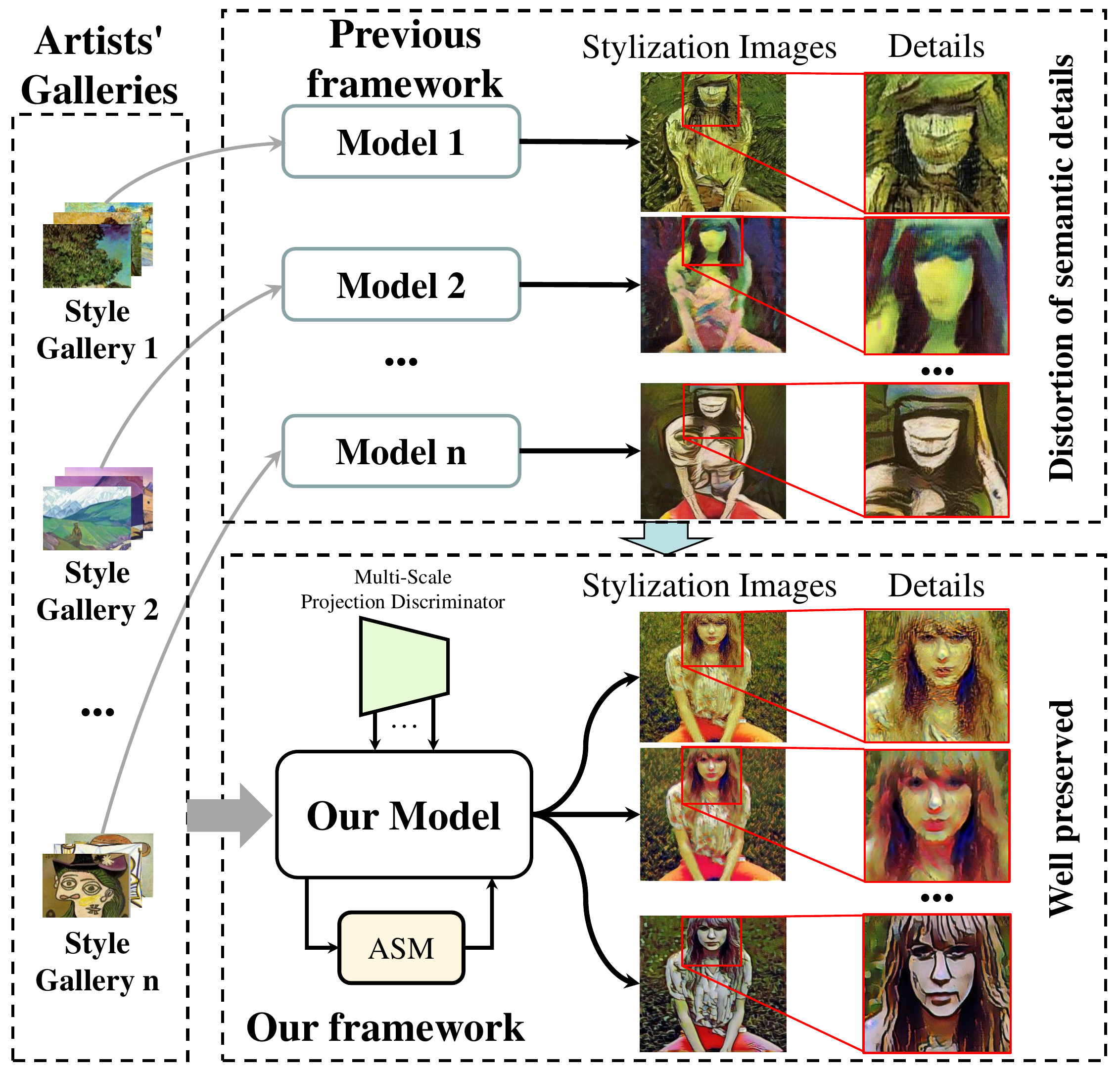}
\end{center}
    \caption{Previous method~\cite{DBLP:conf/eccv/SanakoyeuKLO18} needs separate models for each target style from artists' portfolios. Meanwhile, the semantic details are unrecognizable with excessive distortion. Our model is able to perform multi-artist style transfer within one model, while well preserving semantic information through anisotropic style-stroke controlling.}
\label{motivation}
\end{figure}

Style transfer is a practical technique that transfers a natural photograph into an artistic painting.
Recently, Convolutional Neural Network (CNN) based style transfer approaches~\cite{DBLP:journals/corr/GatysEB15a,DBLP:conf/eccv/SanakoyeuKLO18,DBLP:conf/cvpr/YaoRX0LW19} make significant progress in imitating style texture and tone.
However, a convincing stylization is not just about the imitation of textures, it also needs to choose suitable stroke size according to different semantic regions, \emph{e.g.,} face, background.
Painters never use the same stroke size on the entire painting, where they use thicker strokes in trivial regions(e.g., sky, water surface), and use finer strokes in non-trivial regions(e.g., face, boats) to show more details. Therefore, anisotropic stroke control according to the content is a vital idea to make the style transfer closer to real painting. In addition, from a practical point of view, the style transfer model should have the ability of achieving multiple artistic stylization just via ONE single model.

\begin{figure*}[t]
\begin{center}
\includegraphics[width=1\linewidth]{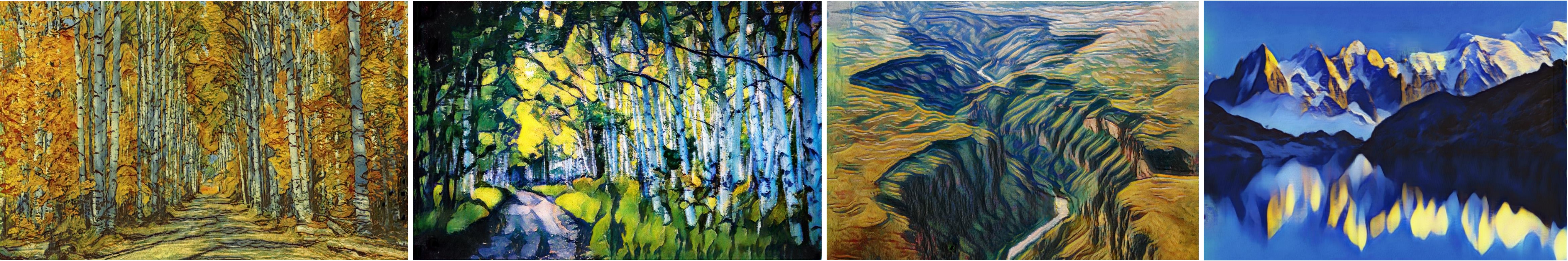}
\end{center}
\caption{Some paintings generated by our ASMA-GAN. From left to right, the corresponding artists are Van Gogh, Samuel, Munch, Nicholas. These results are very similar to oil painting. More high resolution results can be found in suppl.}
\label{attmap}
\end{figure*}

Strokes refer to the fundamental element of the artistic paintings,
artists tend to use different stroke size in various part of a painting.
For example, to make the face more vivid, painters use fine brush strokes to outline facial details, while using thicker brush strokes to draw the background.
Approaches such as~\cite{DBLP:conf/eccv/SanakoyeuKLO18} devote to learning style-stroke control in the style transfer.
However, those frameworks can only adjust the overall brushstroke of the painting without distinction.
Inspired by human perception, recent methods such as~\cite{DBLP:conf/cvpr/YaoRX0LW19, DBLP:journals/corr/abs-1811-12670, DBLP:conf/nips/MejjatiRTCK18} incorporate attention mechanism to achieve the different generation granularity in non-trivial regions and trivial regions.
Yao et al.~\cite{DBLP:conf/cvpr/YaoRX0LW19} follow this scheme and introduce self-attention module which generates a salient map in hidden space, and then adjusts the style-stroke of different regions according to that map.
However, this method suffers from the uncertainty of focus areas and poor effect of highlighting the pixel-wise salient part of semantic content.

Additionally, most contemporary style transfer methods~\cite{DBLP:journals/corr/GatysEB15a,DBLP:journals/corr/ChenS16f,DBLP:conf/iccv/HuangB17,DBLP:conf/eccv/JohnsonAF16,DBLP:conf/icml/UlyanovLVL16,DBLP:conf/cvpr/UlyanovVL17} focus on example guided stylization,
which transfers the style characteristics of the example image onto a target content image.
In this way, those approaches can only achieve the imitation of color and texture of a single painting rather than learning the overall artist style of an artist.
Such learning strategy is completely different from human-artistic creation habits.
The common way for humans to learn to paint the style of an artist is to delve into a set of works of the artist instead of a single piece of artwork.
By analogy, the abstract yet comprehensive style-knowledge should be flexibly modeled from a quantity of artist paintings.
Sanakoyeu et al.~\cite{DBLP:conf/eccv/SanakoyeuKLO18} train their model by a set of certain artist artworks and indeed achieve a substantial improvement in visual quality.
However, this method faces strict limitations, only one single artistic stylization can be performed within one model.
Such defects make this method have serious difficulties in deployment, \emph{e.g.}, one thousand styles need a thousand models to deploy, which is unrealistic.

To address the two problems mentioned above, we propose a framework called \emph{Anisotropic Stroke Multiple Artists GAN, ASMA-GAN}.
To solve the anisotropic control of style-stroke problem, we present the \emph{Anisotropic Stroke Module, ASM}.
Worth mentioning, the size of the receptive field determines the size of the style-stroke~\cite{DBLP:conf/eccv/JingLYFYTS18}.
In contrast to~\cite{DBLP:conf/cvpr/YaoRX0LW19}, our ASM does not explicitly generate the salient map to mark different style-stroke size regions, because this map is troublesome to learn without explicit supervised signals.
In order to dynamically adjust the style-stroke, our AMS integrates features from different scales of receptive fields (equivalent to different strokes) according to the control signal.
In detail, the control signal comes from the deepest hidden features, and the features have the largest receptive field, which will yield the thickest style-stroke.
At the same time, these features have the most abundant semantic information, and this information can guide AMS to distinguish the non-trivial region from trivial region.
Therefore, with the help of ASM, our framework is able to carry out dynamic and variable-grain style transfer.
Unlike existing multi-domain translation issues~\cite{DBLP:journals/corr/abs-1806-02169}, multi-artist style transfer requires the framework to have the ability to discriminate multi-scale texture textones.
In view of this, a \emph{Multi-Scale Projection Discriminator} is proposed to utilize multi-scale characteristic of style to integrally extract style information.
In fact, many paintings contain plenty of micro-structures, which will be lost as the network deepens.
Instead of single scale classification, our discriminator judges the authenticity of the painting by drawing the features of different receptive fields and using them comprehensively.
Furthermore, our discriminator abandons the auxiliary classifier adopted by ACGAN, and uses the projection way~\cite{DBLP:conf/iclr/MiyatoK18} to embed the conditional information into the final output likelihood.
With such a design, our discriminator can effectively encourage the generator to synthesize highly realistic stylized results.
Extensive experimental results well demonstrate the effectiveness and high visual quality achieved by our framework.

\begin{figure*}[t]
\begin{center}
\includegraphics[width=1\linewidth]{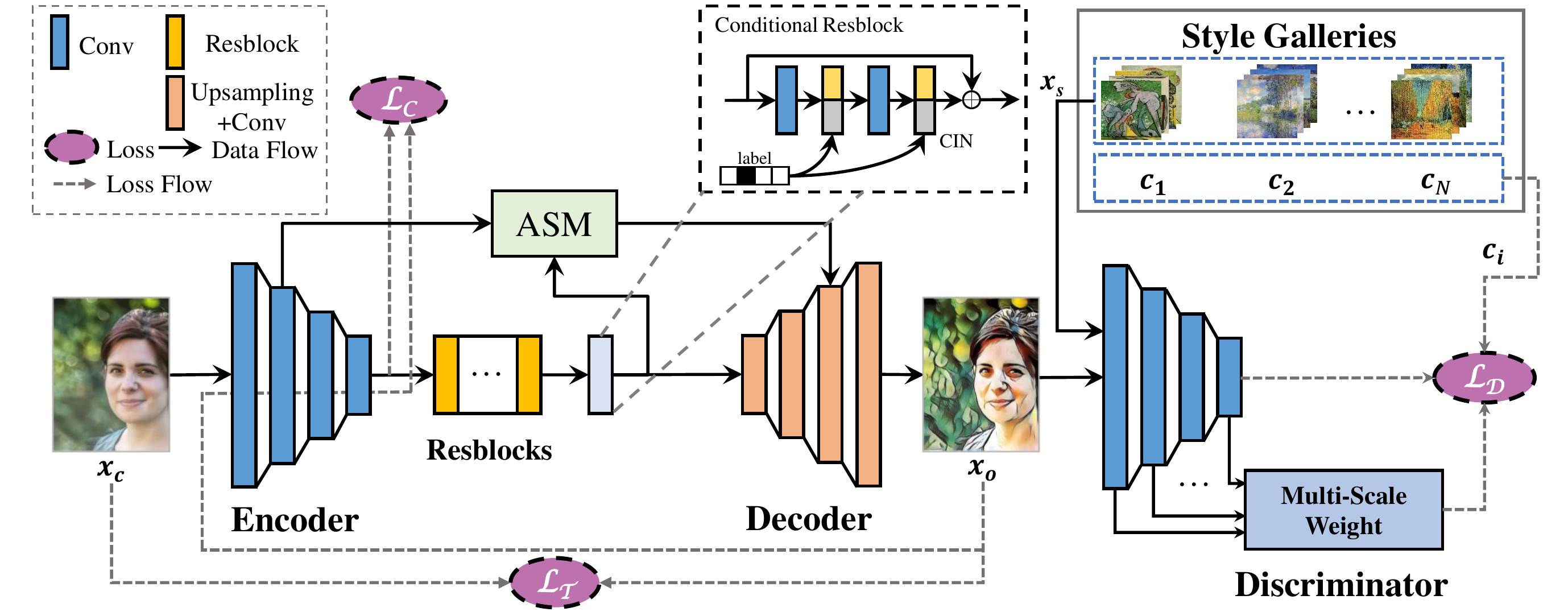}
\end{center}
    \caption{An overview of our framework. We train the ASM and multi-artist style transfer uniformly.
    ASM can dynamically adjust the stroke size of the corresponding region according to the semantic information.
    Our Multi-Scale Projection Discriminator can learn the multi-scale characteristics of oil painting to better guide the training of Generator.
    Each artist's style is trained by portfolio of certain artist instead of single painting.}
\label{framework}
\end{figure*}

\section{Related Work}
{\bf Semantic focus.}
Motivated by the importance of attention in human perception, 
a lot of research efforts have been devoted to the semantic objects within an image~\cite{DBLP:conf/iccv/HeGDG17,DBLP:conf/eccv/ChenZPSA18,DBLP:conf/nips/VaswaniSPUJGKP17,DBLP:conf/cvpr/ChenYWXY16}. 
Methods in the field of image translation can be categorized as two-step and one-step. 
Two-step methods~\cite{DBLP:journals/corr/Champandard16,DBLP:conf/pcm/XiaoZZ18} split models into two separated phases: 
1) acquire semantic mask from a separated segmentation network;
2) process the semantic focal area using the main model. 
To augment the ability of adapting to variations of semantic context,
Luan et al.~\cite{DBLP:conf/pcm/XiaoZZ18} use DilatedNet~\cite{DBLP:journals/pami/ChenPKMY18} to generate image segmentation masks of the inputs and reference images for a set of common labels. 
Two-step methods require an extra pre-training network and millions of labels on semantic context.
Therefore, they are time-consuming and also dilate the structures.
By contrast, one-step models~\cite{DBLP:conf/cvpr/MaFCM18,DBLP:conf/nips/MejjatiRTCK18} achieve semantic focus by incorporating the attention mechanism within the intact model.
Ma et al.~\cite{DBLP:conf/cvpr/MaFCM18} decouple local textures from holistic shapes by attending to local objects of interest through square image regions,
while it results in alteration of background during image translation. 
Mejjati et al.~\cite{DBLP:conf/nips/MejjatiRTCK18} explore an attention network to circumvent the problem. 

{\bf Image Translation.} 
Style transfer is a subfield of image translation where the goal is to learn the mapping between style and content images.
The key issues of style translation are the presentation of style and  the synthesis of image.
Since the success of NST proposed by Gatys et al.~\cite{DBLP:journals/corr/GatysEB15a}, neural representation of image is widely applied in texture synthesis. 
To speed up the style transfer process, several algorithms have been proposed~\cite{DBLP:conf/eccv/JohnsonAF16,DBLP:conf/icml/UlyanovLVL16,DBLP:conf/cvpr/UlyanovVL17},
which  produce stylized results with a forward pass.
To improve the flexibility, models incorporating multiple and arbitrary styles are proposed~\cite{DBLP:journals/corr/ChenS16f,DBLP:conf/iccv/HuangB17,DBLP:conf/nips/LiFYWLY17,DBLP:journals/corr/abs-1805-04103,DBLP:journals/corr/abs-1805-11155}. 
These works synthesize style texture by the representation of style captured from certain artwork rather than the style domain.
Many works achieve domains mapping using generative adversarial networks (GANs) by unpaired images~\cite{DBLP:conf/iccv/ZhuPIE17,DBLP:conf/iccv/YiZTG17,DBLP:conf/icml/KimCKLK17}.
Sannakoyeu et al.~\cite{DBLP:conf/eccv/SanakoyeuKLO18} utilize related style images to train an adversarial discriminator and optimize the generator with content perceptual loss. 
AC-GAN~\cite{DBLP:conf/icml/OdenaOS17} provides class information to generator 
and modifies the learning target of GANs by an auxiliary classifier.
Instead of naively concatenating class information to the input, Projection Discriminator~\cite{DBLP:journals/corr/abs-1802-05637} proposes a specific form of the discriminator, motivated by a commonly occurring family of probabilistic models.
However, it only utilizes the feature of the last layer, which would lose style information when transferring.
Our method composes multi-scale style information.

\section{Methodology}
Our framework learns from artists’ portfolios, instead of one single painting, for novel art creation, \emph{i.e.}, multi-artist stylization with flexible style-stroke size.
To this end, we propose a \textit{ASMA-GAN} framework consisting of the following components:
1) a Conditional Generator $\mathcal{G}$ which efficiently leverages multi-artist style labels to synthesis corresponding stylized painting $x_o$;
2) a novel module called \emph{Anisotropic Stroke Module} which endows the generator the capability to adjust style-stroke~\cite{DBLP:conf/eccv/JingLYFYTS18} size in different region according to the semantic information;
and
3) a \emph{Multi-Scale Projection Discriminator} $\mathcal{D}$, which encourages style consistency through the task of distinguishing artworks of different artists.
Figure.~\ref{framework} illustrates the full pipeline of our approach.

\subsection{The Conditional Generator}
From the perspective of art creation, style should be learned from artists’ portfolios instead of a single painting~\cite{DBLP:conf/eccv/SanakoyeuKLO18}.
Our model benefits from this conception: it is trained by artists’ portfolios. Suppose $\mathbb{I}_i\in \{\mathbb{I}_{Monet},\mathbb{I}_{Picasso},...\}$ denotes an artist's portfolio, $x_s\in \mathbb{I}_i$ denotes an artwork of portfolio $\mathbb{I}_i$.
Given an input content image $x_c$ and a target label $l$, the task is to generate a stylized result $x_o$ using our Conditional Generator.
Instead of unskillfully imitating a single painting, we manage to make use of more general characteristics of a certain artist.
The Conditional Generator consists of four parts: the Encoder, the Resblocks, the conditional Resblock and the Decoder.

{\bf Style Information Injection.}
To generate multi-artist stylized images within a single model, efficient injection of style label information is necessary and crucial.
Previous multi-domain translation method~\cite{DBLP:journals/corr/abs-1711-09020} directly concatenates one-hot label map with the input image or the feature map. 
This approach is only suitable for tasks with similar domains.
Since there are a significant discrepancy between the content domain and the style domain in style transfer task.
Therefore, it is invalid to inject the style information into the network through the direct concatenation.
Another serious flaw is that directly concatenated one-hot vector label is invalid after reflection-padding.
Validity of conditional input depends on whether it would change data distribution in feature space~\cite{DBLP:journals/corr/abs-1806-10050}.
Based on that idea, we design a conditional Resblock that uses Conditional Instance Normalization (CIN)~\cite{DBLP:journals/corr/DumoulinSK16} as the style information injection means.
The structure of the conditional Resblock is shown in Fig.~\ref{framework}.

\begin{figure}[t]
\begin{center}
\includegraphics[width=0.8\linewidth]{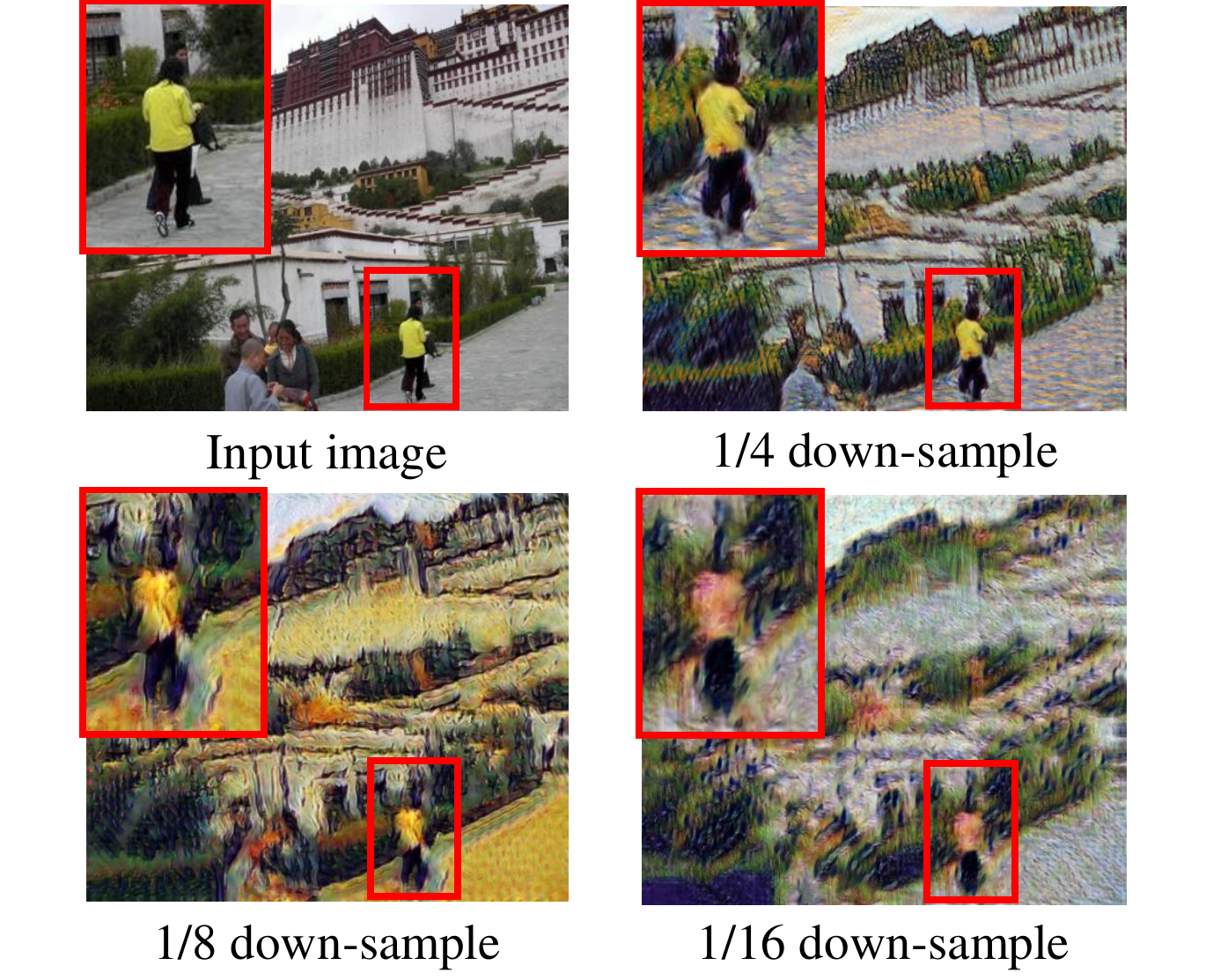}
\end{center}
\caption{Different sizes of style-stroke regard to different down-sample rates.
We put resblocks in places of different down-sample rates between the Encode and the Decoder to show the relationship between down-sample rate and stylization extent.
As shown above, granularity of stylization increases as down-sample rate grows.}
\label{downsample}
\end{figure}

\subsection{The Anisotropic Stroke Module}
When watching an artwork, people are more sensitive to semantic content, such as people, faces and expect them to preserve details with less distortion. 
However, coarse granularity of stylization results in detail distortion. 
For example, fine granularity of stylization distorts subtle objects like edges, while coarse one causes distortion of large scale objects, like cars and human face.
Actually, the granularity is closely related to receptive field of the network, and the larger the receptive field is, the coarser the granularity will be.
In Fig.~\ref{downsample}, it is shown that different down-sample (equivalent to receptive field size) results in stylization with varying granularities.
Since this phenomenon extremely resembles drawing painting with different sizes of painting brushstrokes, it is named style-stroke~\cite{DBLP:conf/eccv/JingLYFYTS18}.
We present our framework to employ the \emph{Anisotropic Stroke Module} to dynamically adjust the style-stroke size according to the semantic information in various region.
Dynamic style-stroke yield pleasing stylized results with meticulous strokes in rich semantics region and rough strokes in remaining region.
The dynamic style-stroke make the stylized results maintain the legibility of the important content (\emph{e.g.,} face, building parts and so on) in the photograph without being severely distorted and losing the meaning of the original picture.

\begin{figure}[t]
\begin{center}
\includegraphics[width=0.9\linewidth]{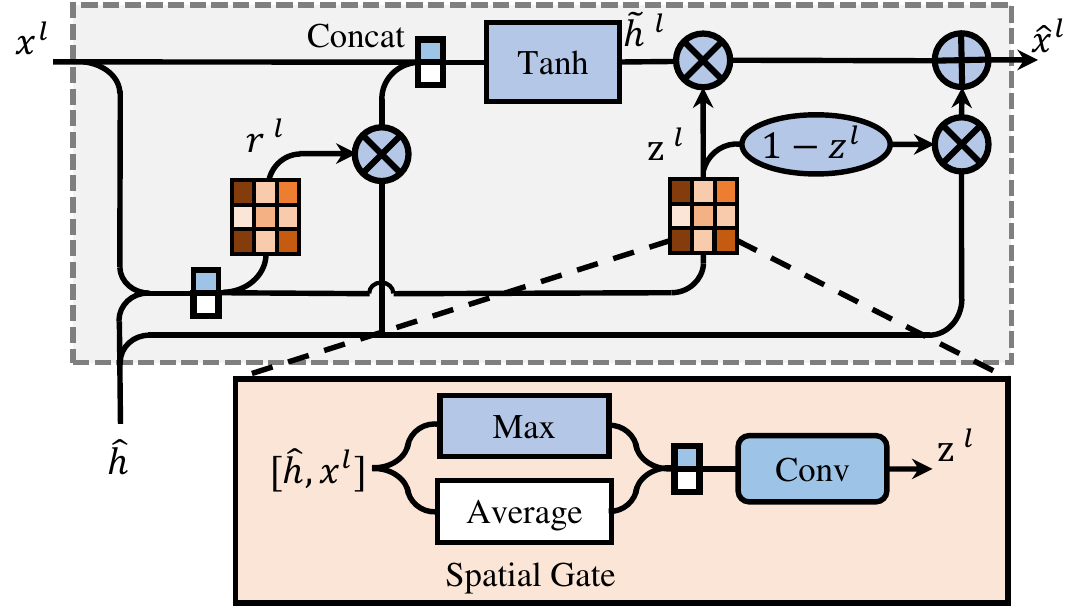}
\end{center}
\caption{Structure of ASM. Our ASM is a variant of GRU with much lighterweight. The box below is our spatial gate structure.
ASM can fuse features of two different scales to achieve dynamic adjustment of style-stroke.}
\label{lstu}
\end{figure}

The detailed structure of ASM is shown in Fig.~\ref{lstu}.
We re-design the reset and update gates with a spatial-wised attention mechanism~\cite{DBLP:conf/eccv/WooPLK18} to be light weighted and still effective in information incorporation.
Instead of removing the gates we choose to lighten them~\cite{DBLP:journals/tetci/RavanelliBOB18} because experiments show removing either of them would lead to a sharp decline in model performance.
Without loss of generality, we choose the ASM located in the $l$-th layer as an analysis example. 
The $l$-th layer feature coming from the encoder side denotes as $\bm{x}^l$.
$\bm{h}$ denotes the feature drawn from the last layer of the bottleneck. It contains rich semantic information to help ASM distinguish important and unimportant regions.
$\bm{h}$ is firstly concatenated with style class information $\bm{c}$ to obtain up-sampled hidden state $\hat{\bm{h}}$.
Then $\hat{\bm{h}}$ and $\bm{x}^l$ are combined to calculate the masks $\bm{r}^l,\bm{z}^l$ for the reset gate and update gate.
$\mathbf{W}_T$, $\mathbf{W}_r$,$\mathbf{W}$ and $\mathbf{W}_z$ represent parameter matrix of transposed convolution, reset gate, merging and update gate. The further process is similar to GRU.
$\hat{\bm{x}}^{l}$ combines two different receptive field features, in other words it blends the style-stroke of two scales.
The equation of gates is shown below,

\begin{equation}
\begin{split}
    \hat{\bm{h}}=\mathbf{W}_T*\left[\bm{h},\bm{c}\right],\quad\quad\quad \\
    \bm{r}^l = \mathbf{W}_r*\left[\hat{\bm{h}},\bm{x}^l\right],\quad\quad\ \ \\
    \tilde{\bm{h}}^l = tanh\left(\mathbf{W}*\left[\bm{r}^l\circ \hat{\bm{h}},\bm{x}^l\right]\right),\\
    \bm{z}^l = \mathbf{W}_z*\left[\hat{\bm{h}},\bm{x}^l\right],\quad\quad\ \ \\
    \hat{\bm{x}}^{l} = \bm{z}^l\circ \tilde{\bm{h}}^l + \left(1-\bm{z}^l\right)\circ \hat{\bm{h}}.\ \ \
\end{split}
\end{equation}

\subsection{The Multi-Scale Projection Discriminator}
\quad The discriminator is the most important component of GANs, and it is trained in the game between the generator and itself.
The Discriminator acts as an oil painting connoisseur in our framework, and its performance directly determines the visual quality of style transfer results.
The existing translation frameworks achieve multi-domain discriminator in the following two ways: 
1) Adding an auxiliary classifier similar to AC-GAN~\cite{DBLP:conf/icml/OdenaOS17}; 2) Using multiple discriminators~\cite{DBLP:conf/accv/YuCYLL18}.
In the first method, the auxiliary classifier works well at low domain variance, but it is difficult to show good performance when the variance is high.
In the second method, GANs are known for its notoriously difficult training, and multiple discriminators make training more unstable.
Takeru Miyato et al.~\cite{DBLP:journals/corr/abs-1802-05637} propose a new conditional generation method for multi-class images synthesis.
This method masterly projects the class label information into the likelihood and shows the state-of-the-art performance in the Imagenet~\cite{ILSVRC15}.
They decompose the adversarial likelihood to the sum of two components:

\begin{equation}
    \mathcal{D}(\bm{x_o},\bm{c})=\bm{c}^T\bm{V}\phi(\bm{x_o})+\psi(\phi(\bm{x_o})),
    \label{proj-d}
\end{equation}
where $\bm{V}$ denotes an embedding matrix.
$\phi(\cdot)$ represents the input to the last layer of the convolution network part of $\mathcal{D}$.
$\psi(\cdot)$ is the FC layers to scale the output.

\begin{figure}
\begin{center}
\includegraphics[width=0.9\linewidth]{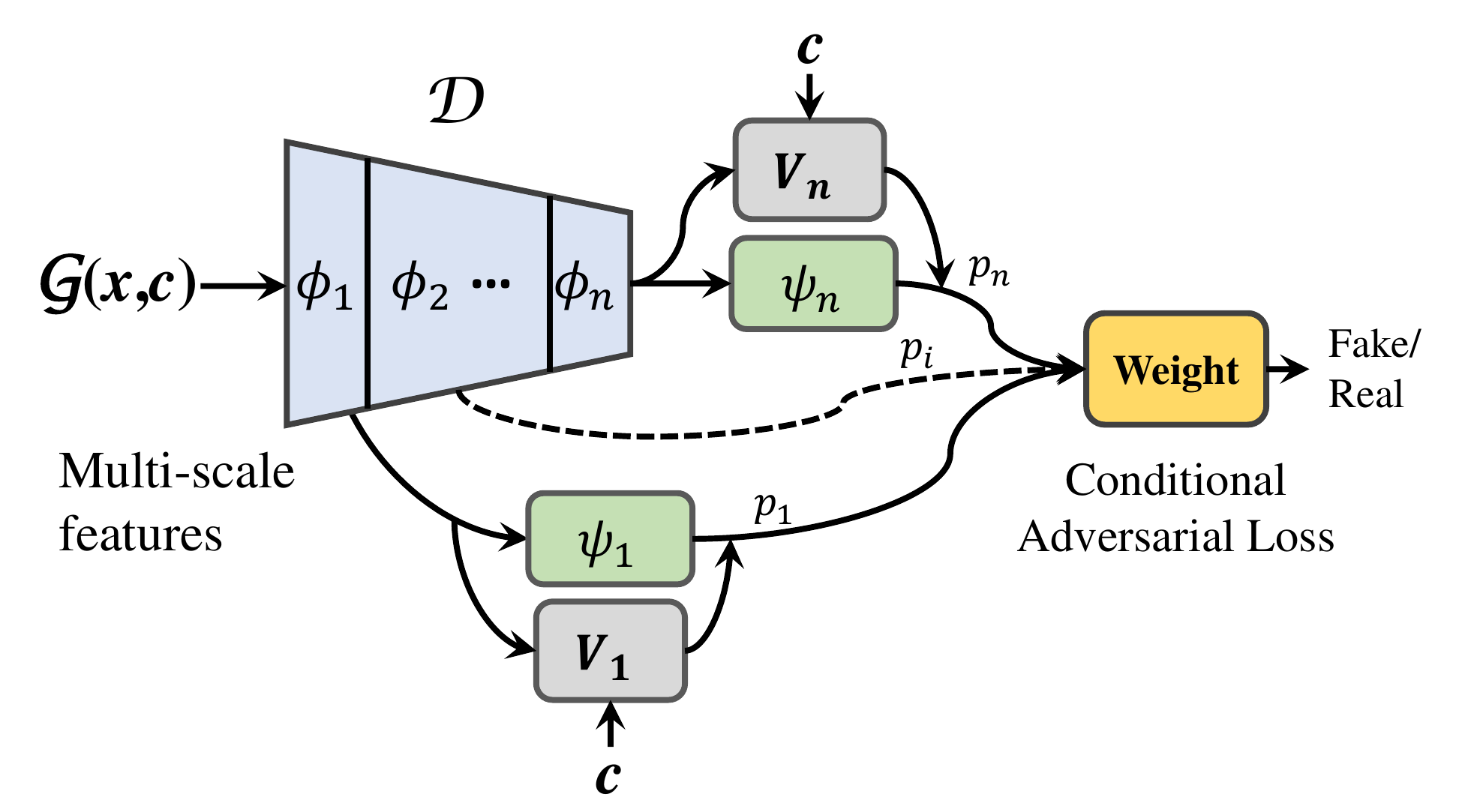}
\end{center}
\caption{Unlike the Projection Discriminator~\cite{DBLP:conf/iclr/MiyatoK18}, our model comprehensively uses the features of different scales,
which greatly strengthens the discriminator to recognize the stroke textons of different scales in the painting.}
\label{multi-scale-d}
\end{figure}

Inspired by PatchGAN~\cite{DBLP:conf/cvpr/IsolaZZE17}, we design a novel discriminator, called \emph{Multi-Scale Projection Discriminator}, for extracting the multi-scale characteristics of oil paintings while achieving the multi-artist style transfer.
Worth mentioning, the multi-scale character of artist style is exceedingly commonplace in oil paintings.
For example, Van Gogh's paintings, his paintings have unique overall color and texture features, and are characterized by neatly arranged short strokes.
The most obvious difference between our Multi-Scale Projection Discriminator and~\cite{DBLP:conf/cvpr/IsolaZZE17} is that we extend the discriminator into a multi-scale one that can fuse multiple scale features.
It can capture the style characteristics of an artist's painting set from strokes to color schemes.
The structure of our discriminator is shown in Fig.~\ref{multi-scale-d}.
Additionally, as the training process of GANs is extremely unstable, we apply the Spectral Normalization (SN)~\cite{DBLP:journals/corr/abs-1802-05957} in the Multi-Scale Projection Discriminator, which is able to force the weights in discriminator to regularize the Lipschitz constant yielding a stable training process.
The mathematical expression of our Multi-Scale Projection Discriminator is shown below:

\begin{equation}
    \mathcal{D}(\bm{x_o},\bm{c})=\sum_{i=0}^{N}w_i\cdot\left[\bm{c}^T\bm{V}_i\phi_i(\bm{x_o})+\psi_i(\phi_i(\bm{x_o}))\right],
    \label{our proj-d}
\end{equation}
where $N$ indicates the total number of feature scales of the Multi-Scale Projection Discriminator, $w_i$ denotes weighting factor of $i$-th output likelihood.

\subsection{Objective Function}
{\bf Perceptual Loss.}
Generator should try to ensure the semantic consistency of content while stylization. 
Most of the existing translation networks~\cite{DBLP:conf/eccv/JohnsonAF16,DBLP:journals/corr/ChenS16f,DBLP:journals/corr/UlyanovVL16,DBLP:conf/eccv/JingLYFYTS18,DBLP:conf/cvpr/YaoRX0LW19} use the pre-trained VGG model on Imagenet as the calculation function for perceptual loss~\cite{DBLP:conf/eccv/JohnsonAF16}. 
However, when painting, the artist thinks about the content of the painting from an artistic point of view rather than the classification. 
Under the constraints of such perceptual loss, the generator can not realize the artistic reconstruction of the content details. 
Inspired by the style-aware content loss~\cite{DBLP:conf/eccv/SanakoyeuKLO18}, we measure the similarity in content between input image $\bm{x_c}$ and stylized $\bm{x}_{o} = \mathcal{G}(\bm{x_c},\bm{c})$ by a style-aware content loss. 
The loss directly uses the Encoder of the Generator instead of the pre-trained VGG model as the calculation function for perceptual loss, which makes the Encoder tend to retain the semantic region related to style. Therefore, our generator achieves better style transfer performance, but the content consistency is drastically reduced. This problem is exactly what our \emph{ASM} had solved.
Transform loss~\cite{DBLP:conf/eccv/SanakoyeuKLO18} $\mathcal{L}_T$ is introduced, as the extra signal, which initializes training and boosts the learning of the primary latent space:
\begin{equation}
    \mathcal{L}_C=\mathbb{E}_{\bm{x_c}}\left[\left\Arrowvert \mathcal{E}(\bm{x_c})-\mathcal{E}(\mathcal{G}(\bm{x_c},\bm{c})) \right\Arrowvert_{\ell_1}\right] ,
    \label{content loss}
\end{equation}
\begin{equation}
    \mathcal{L}_\mathcal{T}= \mathbb{E}_{\bm{x_c}}\left[\frac{1}{CHW}\Arrowvert \mathcal{T}(\bm{x_c})-\mathcal{T}(\mathcal{G}(\bm{x_c},\bm{c})) \Arrowvert_{\ell_1}\right] , \label{transform loss}
\end{equation}
where $\mathcal{E}$ is the Encoder of $\mathcal{G}$, $\mathcal{T}$ denotes a pooling layer, and $C$,$H$,$W$ respectively represent channels, height, width of $\mathcal{T}(\cdot)$. $\Arrowvert \cdot \Arrowvert_{\ell_1}$ denotes $\mathcal{L}_1$ loss. Experiments show that compared to perceptual loss, training with sytle-aware loss can achieve better saturation in the stylized image.

{\bf Adversarial Loss.}
At the beginning of the training process, the stylization results are almost the same as the photographs.
In order to speed up the discriminator to learn to distinguish between paintings and photographs, we also add photographs as the fake samples to the training of discriminator.
Furthermore, we introduce the hinge loss~\cite{DBLP:journals/corr/LimY17} instead of WGAN loss~\cite{DBLP:journals/corr/ArjovskyCB17} as the standard adversarial loss.
Our standard adversarial loss is shown below:
\begin{align}
\begin{split}
    \mathcal{L}_\mathcal{D}=&\ \mathbb{E}_{\bm{c}}\left[\mathbb{E}_{\bm{x_s}}\left[max(0,1-\mathcal{D}(\bm{x_s},\bm{c}))\right]\right]\\
    &\quad +\mathbb{E}_{\bm{c}}\left[\mathbb{E}_{\bm{x_c}}\left[max(0,1+\mathcal{D}(\bm{x_c},\bm{c}))\right]\right]\\
    &\quad \quad +\mathbb{E}_{\bm{c}}\left[\mathbb{E}_{\bm{x_c}}\left[max(0,1+\mathcal{D}(\Hat{\mathcal{G}}(\bm{x_c},\bm{c}),\bm{c}))\right]\right] ,
\end{split}
    \label{LD}
\end{align}
\begin{align}
    \mathcal{L}_\mathcal{G}=-\mathbb{E}_{\bm{c}}\left[\mathbb{E}_{\bm{x_c}}\left[\Hat{\mathcal{D}}(\mathcal{G}(\bm{x_c},\bm{c}),\bm{c})\right]\right],
    \label{LG}
\end{align}
where $\Hat{\mathcal{G}}$ and $\Hat{\mathcal{D}}$ indicates that the corresponding model parameters are fixed and no training.

{\bf Overall Loss.}
To summarize, the full objective of our model is:
\begin{equation}
\begin{split}
    \mathcal{L}(\mathcal{A},\mathcal{G},\mathcal{D})=\mathcal{L}_\mathcal{G}+\mathcal{L}_\mathcal{D}+\lambda_C\mathcal{L}_C +\lambda_\mathcal{T}\mathcal{L}_\mathcal{T},
    \label{total loss}
\end{split}
\end{equation}
where $\mathcal{A}$ indicates the ASM. The weight coefficients: $\lambda_C$, $\lambda_\mathcal{T}$ are mainly to balance the magnitude of different loss. We set $\lambda_C=90,\lambda_\mathcal{T}=100$ respectively. 
\begin{figure*}[t]
\begin{center}
\includegraphics[width=1\linewidth]{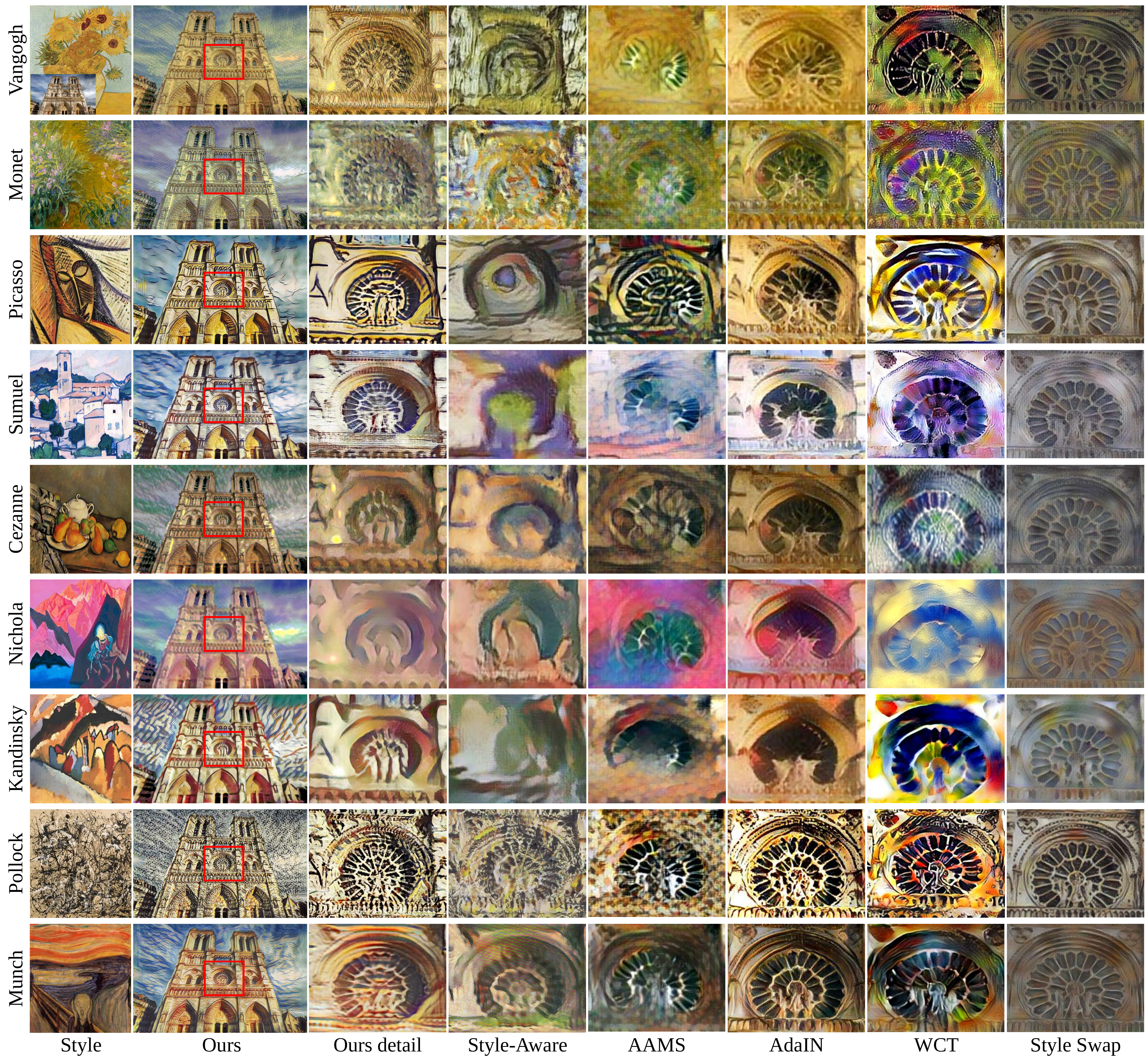}
\end{center}
\caption{Results produced for different artist style from different style transfer methods, with a fixed content image. We evaluate the zoomed in cut-out of the same region in all results to compare the variation across different styles and details stylization effect. More high resolution results can be found in suppl.}
\label{exp1}
\end{figure*}

\begin{table}[!htbp]
\begin{center}
\caption{User studies scores ($mean$) of different methods, in terms of the style transfer effect and anisotropic semantic preserving effect. \emph{Preference Score} is the final score, it is the average of the two scores.}
\begin{tabular}{lc|cc}
\hline
Method&\emph{\makecell[c]{Preference \\ Score}}&\emph{\makecell[c]{Style \\ Deception \\ Score}}&\emph{\makecell[c]{Semantic \\  Retention\\  Score}}\\
\hline
\text{Ours}&  $\bm{8.00}$  &  $7.9$ & $\bm{8.1}$\\
Style-Aware~\cite{DBLP:conf/eccv/SanakoyeuKLO18}&  $7.00$  & $\bm{8.5}$ & $5.5$\\
Style Swap~\cite{DBLP:journals/corr/ChenS16f}&  $5.15$   & $2.7$ & $7.6$\\
WCT~\cite{DBLP:conf/nips/LiFYWLY17}&  $5.95$   & $6.1$ & $5.8$\\
AdaIN~\cite{DBLP:conf/iccv/HuangB17}&  $6.15$   & $6.4$ & $5.9$\\
AAMS~\cite{DBLP:conf/cvpr/YaoRX0LW19}&  $6.25$ & $4.6$ & $7.9$ \\
\hline
\end{tabular}
\label{score1}
\vspace{-5mm}
\end{center}
\end{table}

\begin{figure*}[t]
\begin{center}
\includegraphics[width=1\linewidth]{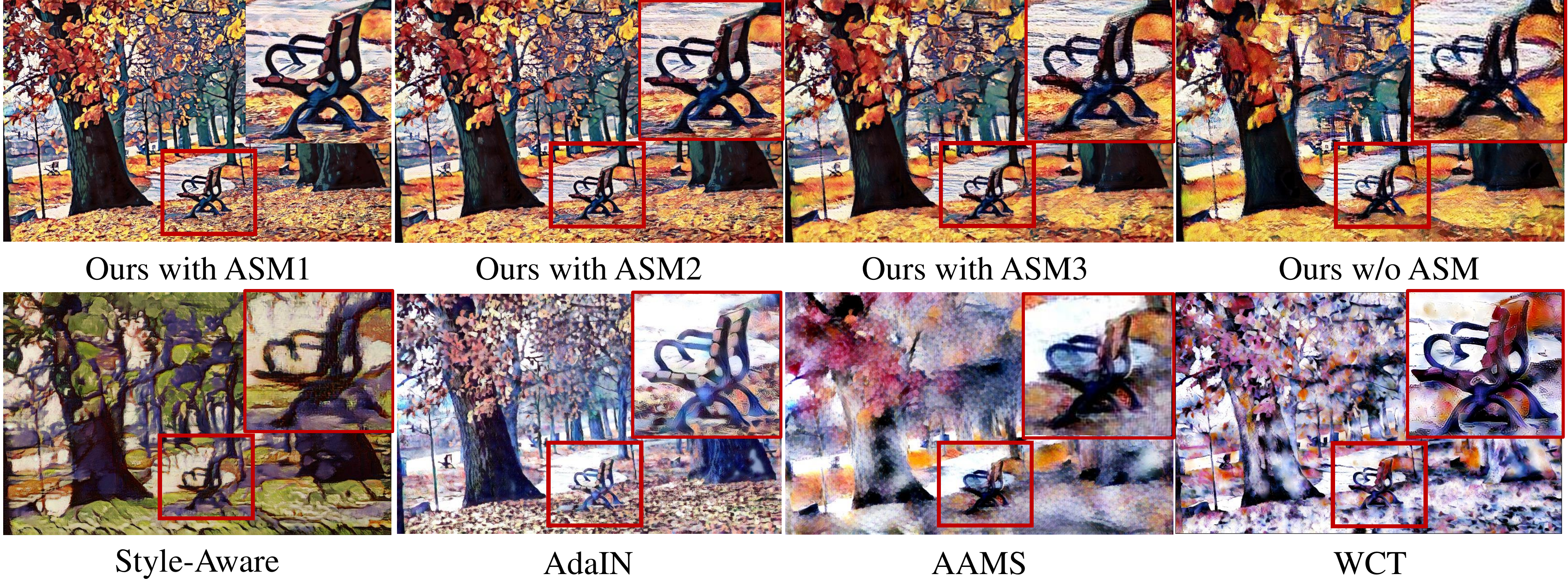}
\end{center}
\caption{Comparison of anisotropic semantic preserving effect from Style-Aware, AAMS, AdaIN, WCT and ours. ASM1, ASM2, ASM3 indicate that ASM is placed in different layers of Generator.}
\label{attmap}
\end{figure*}

\begin{figure}[t]
\begin{center}
\includegraphics[width=0.8\linewidth]{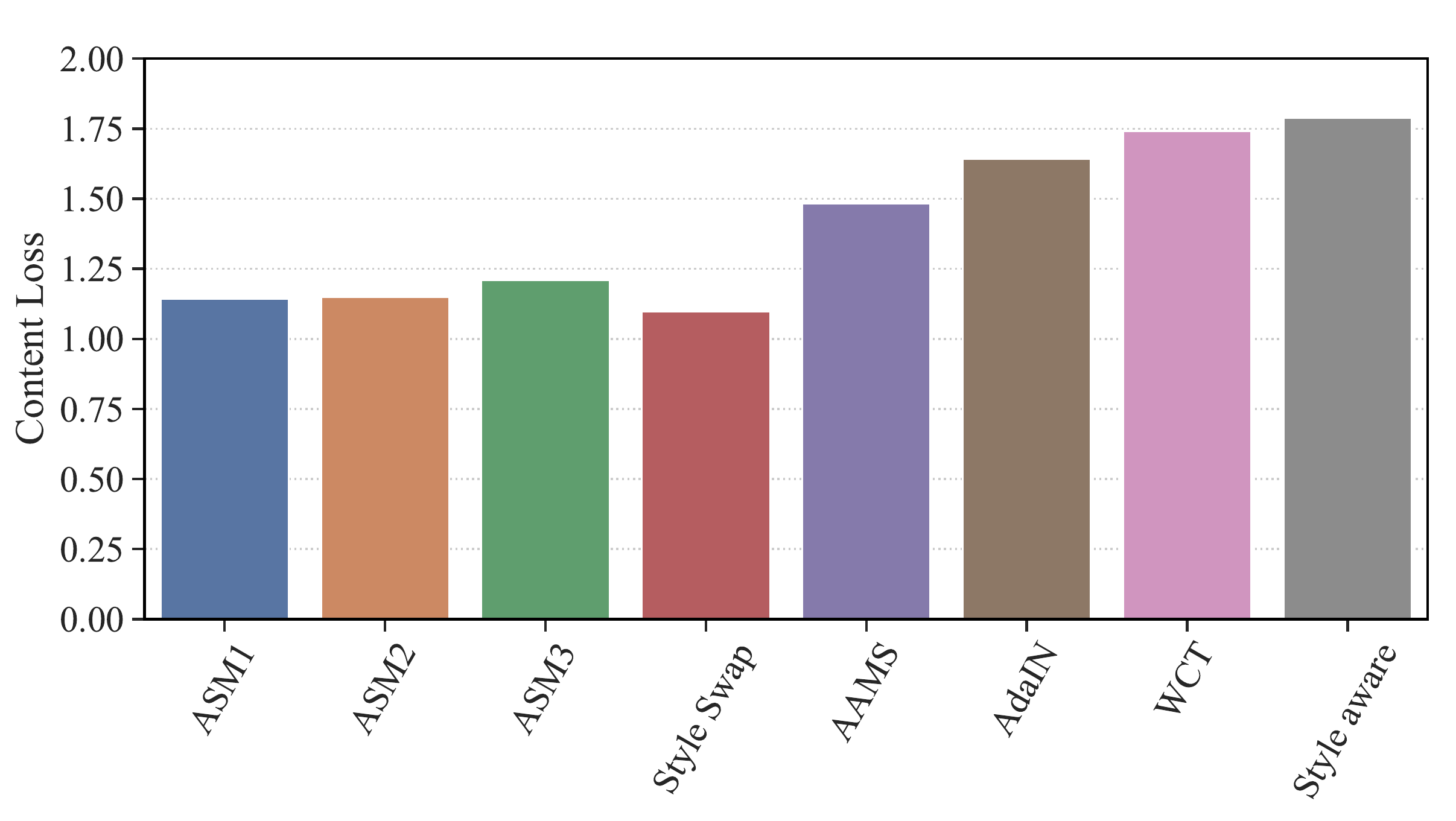}
\end{center}
\caption{Semantic Retention Ratio histogram of different methods. The higher the score, the worse the semantic retention of the corresponding method.}
\label{Ablation1}
\end{figure}

\section{Experiments}
\subsection{Implementation details}
\textbf{Structure details.} As mentioned above, the framework consists of the Conditional Generator,  the Anisotropic Stroke Module and the Multi-Scale Projection Discriminator.
The Conditional Generator contains three blocks: the Encoder, the Resblocks and the Decoder. The Encoder has 1 conv3-stride-1 and 4 conv3-stride-2, where each convolution layer is followed by an IN~\cite{DBLP:journals/corr/UlyanovVL16} and a LeakyRelu. 
5 residual layers~\cite{DBLP:conf/cvpr/HeZRS16} and a conditional Resblock are connected in series to form the Resblocks.
The Decoder is composed of 4 upsampling-conv3 layers and 1 conv3-stride-1 colorization layer. 
The backbone network of the Multi-Scale Projection Discriminator is a fully convolutional network with 6 conv5-stride2-SN-LeakyRelu blocks.
More details can be found in suppl.

\textbf{Training Data.} The training data consists of two parts: the content images are sampled from Places365~\cite{DBLP:journals/pami/ZhouLKO018} and the artistic style portfolios are collected from the Wiki Art dataset.
Optimizer is Adam optimizer~\cite{DBLP:journals/corr/KingmaB14} and learning rate is set as 0.0001.
\subsection{Stylization Assessment}
In order to assess the quality of the stylization results of our framework, we propose two Quantitative metrics: \emph{Semantic Retention Ratio (SRR)} and \emph{Stylization Accuracy}.
Actually, style is a relatively abstract concept, it is difficult to use quantitative metrics for comprehensive measurement.
Based on this fact, we introduced two types of user studies, \emph{Style Deception Score, Semantic Retention Score}, with reference to ~\cite{DBLP:conf/iccv/KotovenkoSLO19,DBLP:conf/cvpr/KotovenkoSMLO19,DBLP:conf/eccv/SanakoyeuKLO18} to perceptually evaluate the effectiveness of our algorithm.

\textbf{Semantic Retention Ratio.}
This metric is designed to measure the degree of retention of the input image semantics after stylization.
In fact, the main task of our ASM is to retain discriminative semantic information.
Therefore, SRR can accurately and quantitatively evaluate the effectiveness of ASM.
The mathematical form of our SRR is as follows:
\begin{equation}
    SSR= \mathbb{E}_{\Delta x}\left[\frac{1}{n^2} \sum_{i,j}\left| \frac{D_{ij}^{\bm{x}_o}}{\sum_{i}D^{\bm{x}_o}_{ij}} - \frac{D_{ij}^{\bm{x}_c}}{\sum_{i}D^{\bm{x}_c}_{ij}} \right|\right],
    \label{total loss}
\end{equation}
where $D^{\cdot}_{ij}$ denotes the patch of corresponding image, $n$ is the total number of patches.

\textbf{Stylization Accuracy.}
We designed the Stylization Accuracy to evaluate the performance of our Multi artist style transfer.
This metric is measured by an artistic style classifier that is isomorphic to our discriminator.
To avoid overfitting, we collected 11 artists' paintings to generate more than $10^5$ style patches to train our style classifier.
Finally, the classifier achieves a \emph{$91.9\%$} average accuracy on the paintings dataset.

\subsection{Qualitative analysis}
\textbf{Style transfer results.} We evaluate our approach with five state-of-the-art methods: AdaIN~\cite{DBLP:conf/iccv/HuangB17}, Style-Aware~\cite{DBLP:conf/eccv/SanakoyeuKLO18}, AAMS~\cite{DBLP:conf/cvpr/YaoRX0LW19}, Style Swap~\cite{DBLP:journals/corr/ChenS16f} and WCT~\cite{DBLP:conf/nips/LiFYWLY17}.
Noting that style aware includes two subsequent works~\cite{DBLP:conf/iccv/KotovenkoSLO19,DBLP:conf/cvpr/KotovenkoSMLO19} and their effects are not much different.
For simplicity, we use style aware to represent this type of method.
We pick the most representative paintings from training style portfolios as the style images to represent the portfolio. 
By comparing the zoomed in cut-outs in Fig.~\ref{exp1}, our method shows much more stunning effect than other competitors. 
The details of the semantic contents ( e.g the clock on the clock tower) are accurately preserved and the characteristic of target styles are vividly maintained. 
Although results of Style-Aware have the most prominent style characteristics, but details in the results are unrecognizable with excessive distortion, caused by the coarse granularity.
It can be seen that the example based methods (WCT, AdaIN, Style Swap, AAMS) cannot effectively learn the characteristics of style.
Their results only show strong tone changes and irregular details distortion.
And these changes strongly depend on the example image.

\begin{figure}[t]
    \begin{center}
    \includegraphics[width=1\linewidth]{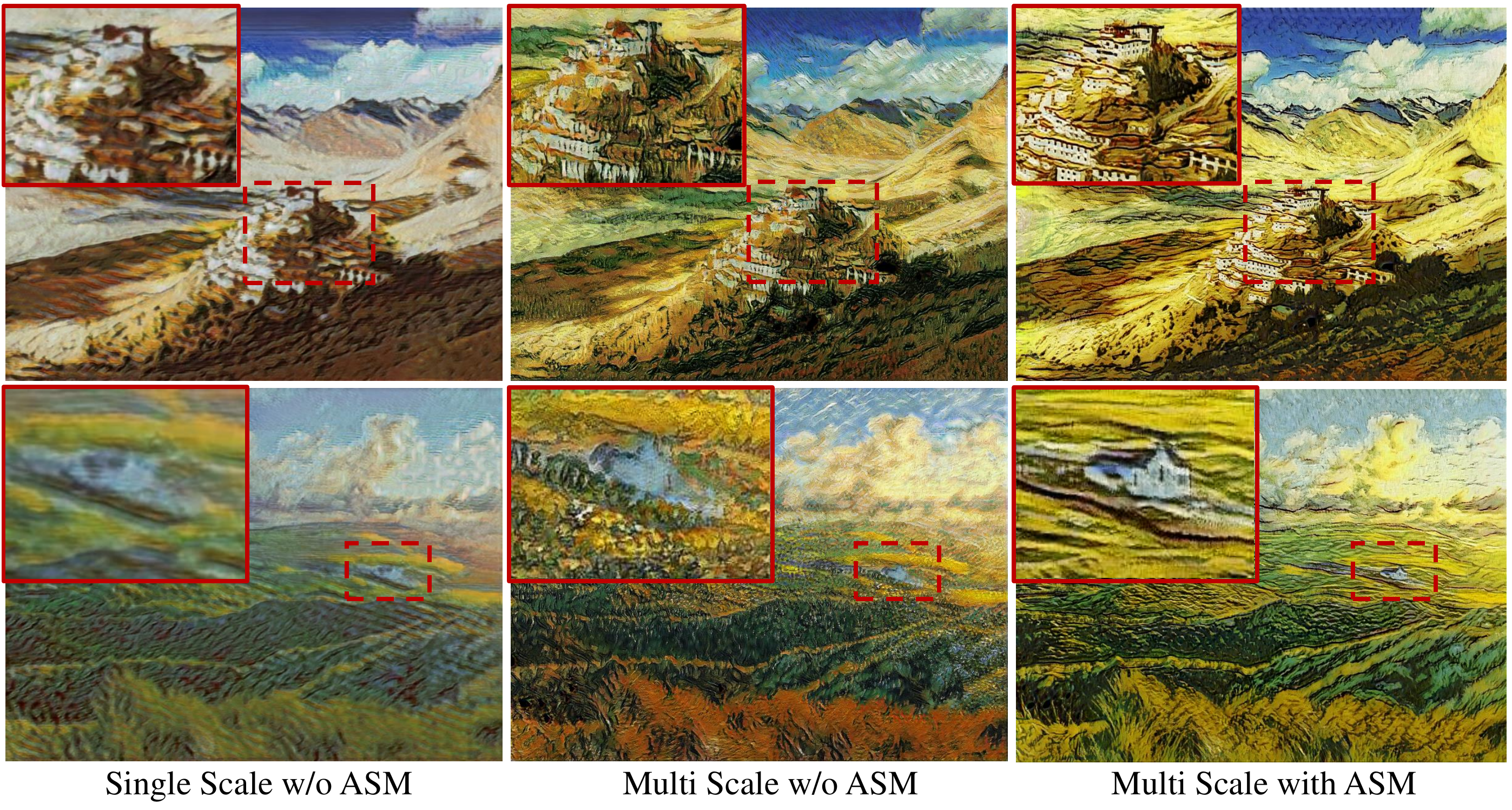}
    \end{center}
    \caption{Comparison of the Multi-Scale Projection Discriminator and the Single-Scale Projection Discriminator.}
    \label{Ablation2}
    \end{figure}
    
    \subsection{Ablation Studies}
    The indispensable two components of our framework are the ASM and the Multi-Scale Projection Discriminator. 
    We study the effectiveness of these two modules by individually removing them. 
    First, by removing the ASM, our model degenerates into a multi-artist style transfer framework.
    The degraded model is trained using the same conditions as described in implementation details section, and the transfer results are shown in Fig.~\ref{attmap}. Compared to other methods, our approach retains the most complete details.
    In addition, we place ASM on different layers in Generator, and use ASM1, ASM2, and ASM3 to represent $1/2$ downsampling, $1/4$ downsampling, and $1/8$ downsampling positions, respectively.
    It can be seen from the figure that the details of the bench gradually become blurred, which indicates that ASM can automatically integrate features from different scales so that our model can automatically use different stylized strokes in different areas.
    
    Then, we remove the multi-scale style learning module from the Multi-Scale Projection Discriminator.
    Using this Single-Scale Projection Discriminator, we train the model on the portfolios mentioned above. The results are shown in Fig.~\ref{Ablation2}.
    In the figure, the first column is the stylization result of the Single-Scale Discriminator.
    Although the results contain some characteristics of the corresponding style, while its details are over-smooth.
    It is because the discriminator lacks the use of shallow features, thus losing the ability to judge details.
    The results confirm that multi-scale style learning module is an indispensable part of our framework.
    The above two ablation experiments demonstrate the necessity and effectiveness of the two modules.
    
    \subsection{Quantitative analysis}
    \textbf{User study.}
    We use 200 groups of images, each consists of the input content image, the target style set and 5 results from~\cite{DBLP:conf/iccv/HuangB17,DBLP:journals/corr/ChenS16f,DBLP:conf/eccv/SanakoyeuKLO18,DBLP:conf/cvpr/YaoRX0LW19} and ours. 
    The user studies include two parts in terms of the style transfer effect and anisotropic semantic preserving effect.
    In the first study, the participants are asked to score the results by the degree of style reduction, from 0 to 10 (10 is the best, 0 is the worst), i.e. \textit{Style Deception Score}. 
    In the second study, the participants score the results by the degree of detail retention of the semantic content, i.e. \textit{Semantic Retention Score}.
    We calculate the mean value of two scores of each method over all participants, as shown in Tab.~\ref{score1}. 
    The studies show that our multiple artists stylized results achieve approximate equivalent effect as the stylized results in~\cite{DBLP:conf/eccv/SanakoyeuKLO18}, which is better than other methods. 
    And our method outperforms others on the semantic detail retention.
    
    \textbf{Content Discrepancy.}
    We carefully picked 200 pictures with abundant semantic information (\emph{e.g.}, portraits, buildings, etc.) from the Place365 to form the benchmark.
    We estimate SSR based on this benchmark.
    It can be seen from Fig.~\ref{Ablation1} that style-swap has the highest score, but its stylization effect is too poor. Our method can achieve good semantic retention no matter where ASM is placed.

\begin{figure}[t]
    \begin{center}
    \includegraphics[width=1\linewidth]{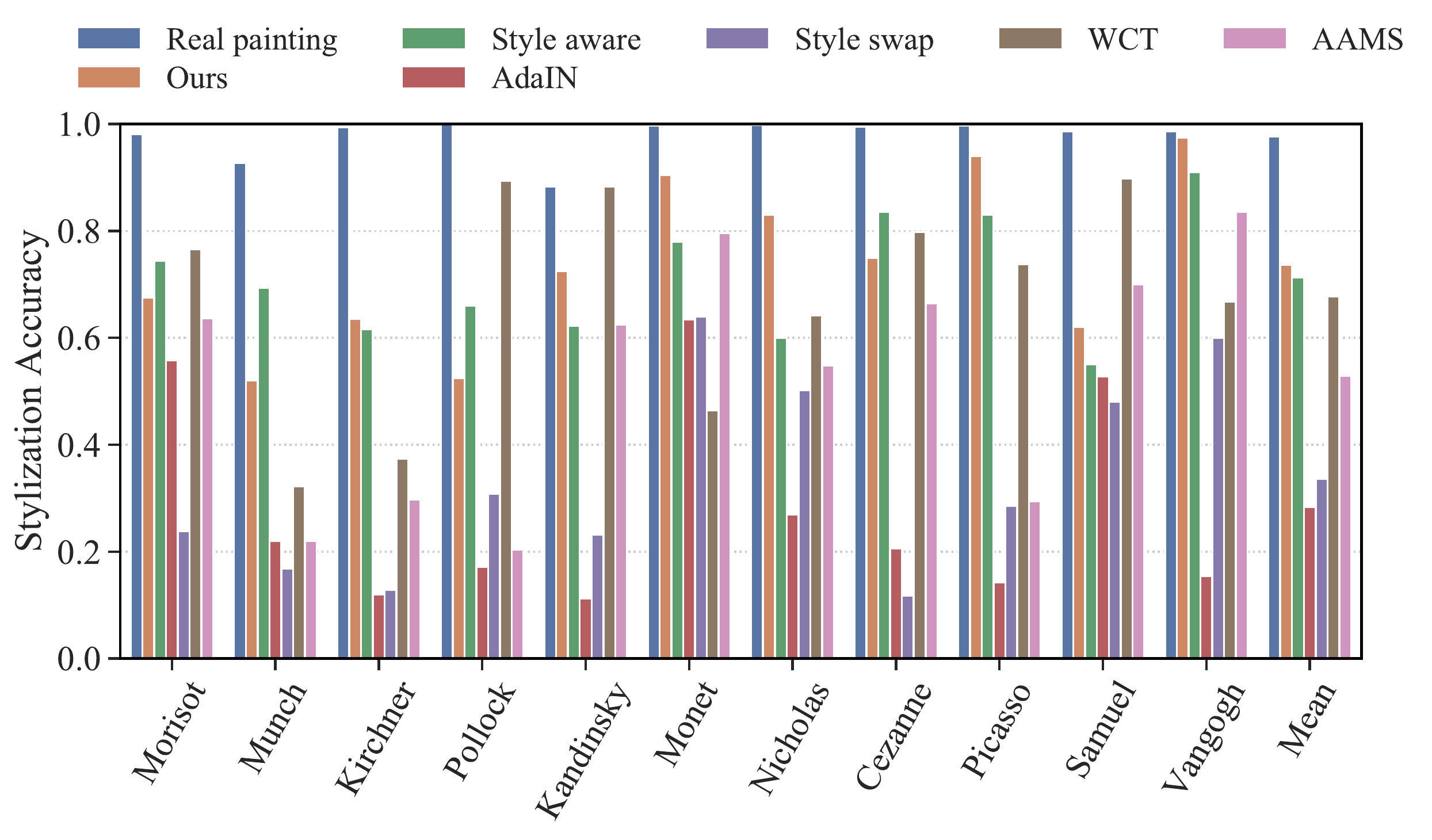}
    \end{center}
    \caption{Our method achieves the best average classification accuracy.}
    \label{styleacc}
    \end{figure}

    \textbf{Style Accuracy.}
    We generate 200 result images for each artist's style, and measure the Style Accuracy of the stylization by sending these result images to the style classifier.
    The higher the accuracy of the classification result, the closer the class is to the corresponding painting style.
    However, style is a perceptual concept, so this indicator can only be used as a reference.
    The classification results are shown in Fig.~\ref{styleacc}.
    
    \section{Conclusion}
    In this paper, we propose a novel Multi-Scale Projection Discriminator, which overcomes the limitation of Single-Scale Projection Discriminator and gives our discriminator ability to deal with the multi-scale characteristics of style.
    Moreover, the ASM is able to dynamically adjust the strokes based on the semantic information of the picture.
    With the help of ASM, our model can retain the vital semantic information of the picture while transferring the style.
    Experimental results demonstrate the effectiveness and delicate visual performance of our method.

\begin{acks}
This work was supported by National Science Foundation of China (61976137, U1611461, U19B2035) and STCSM(18DZ1112300).
\end{acks}

\bibliographystyle{ACM-Reference-Format}
\bibliography{1964}

\appendix

\section{Content}

In this supplementary material, we elaborate on the specific structures of the networks used and the training details. We also provide more generated paintings of different artists style to show the effectiveness of our method.

The contents are given in the following sequence:
\begin{enumerate}
    \item The network structure of ASMA-GAN. This section contains two sub-sections: 1) detailed network structure; 2) The training strategy of generator and discriminator.
    \item The paintings generated by the single network we proposed under the style of eleven painters. In this section, we provide the generation effect of more high-definition and large-size images.
\end{enumerate}
Images are best viewed in color and zoomed in.
Our source code will be made available soon in Github. And more generated pictures are in the compressed package.
 

\section{Implement Details of ASMA-GAN}

\subsection{Detailed Network Structure}
\begin{figure}
\begin{center}
\includegraphics[width=1.0\linewidth]{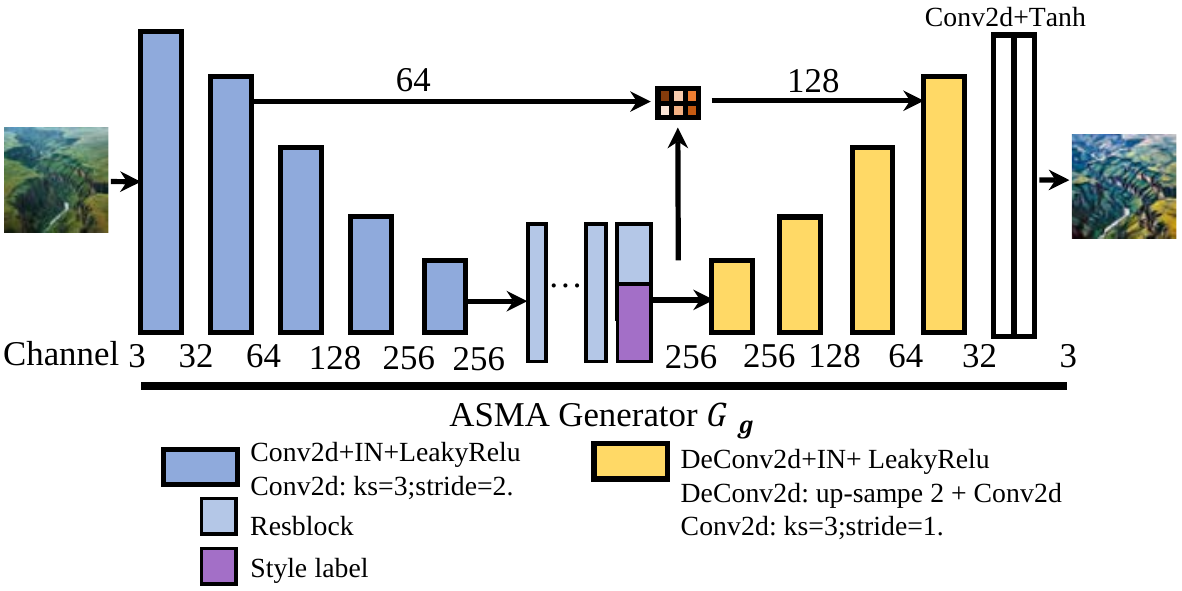}
\end{center}
\vspace{-3 mm}
  \caption{Network structure and parameter details of ASMA-GAN generator, where ks means the kernel size of convolution layers.}
\label{Generator}
\end{figure}

\begin{figure}
\begin{center}
\includegraphics[width=1.0\linewidth]{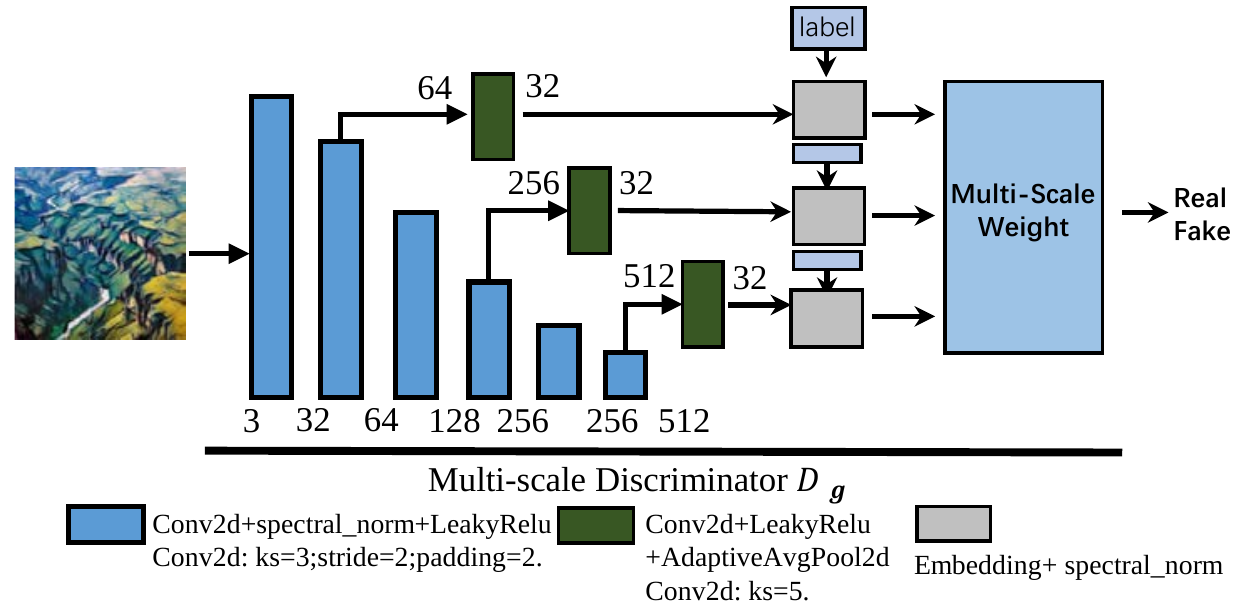}
\end{center}
\vspace{-3 mm}
  \caption{Network structure and parameter details of ASMA-GAN discriminator.}
\label{Disciminator}
\end{figure}

The structure of ASMA-GAN (mentioned in our paper) is given. The structure of the generator is shown in Fig.~\ref{Generator}. The structure of the discriminator is shown in Fig.~\ref{Disciminator}.

\subsection{Training Strategy.}
The three components of our framework are jointly trained.
As the SNs make the discriminator training process become more difficult.
We alternately train the Multi-Scale Projection Discriminator and the Conditional Generator, where 3 times for the Multi-Scale Projection Discriminator and once for the Conditionals Generator.
Due to memory limitations, we start training the framework at the resolution of $256^2$.
After the $10^5$ steps training, we increase the resolution to $512^2$ and continue training $10^5$ steps.
Finally, we increase the resolution to $768^2$.

\section{Results of Different Style Transfer Methods}

In the text, due to space limitations, we only show the details of the results of different style transfer methods. In Fig.~\ref{Portrain_picasso}, we provide the global result images of different methods on the same image of Munch's style.

\begin{figure*}
\begin{center}
\includegraphics[width=1.0\linewidth]{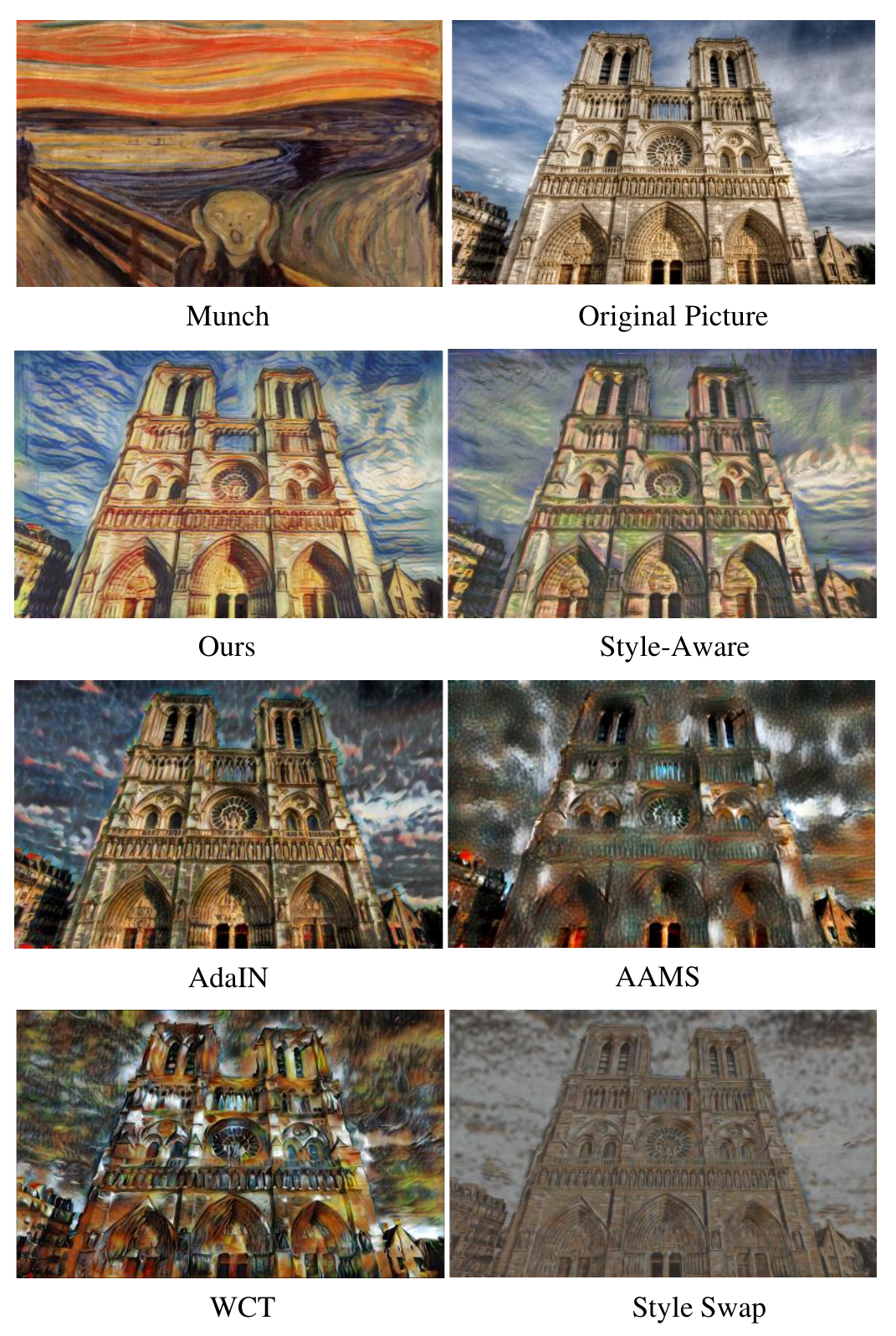}
\end{center}
\vspace{-3 mm}
  \caption{Results produced for different artist style from different style transfer methods.}
\label{com_re}
\end{figure*}
\section{Extra Results of ASMA-GAN}

\begin{figure*}
\begin{center}
\includegraphics[width=1.0\linewidth]{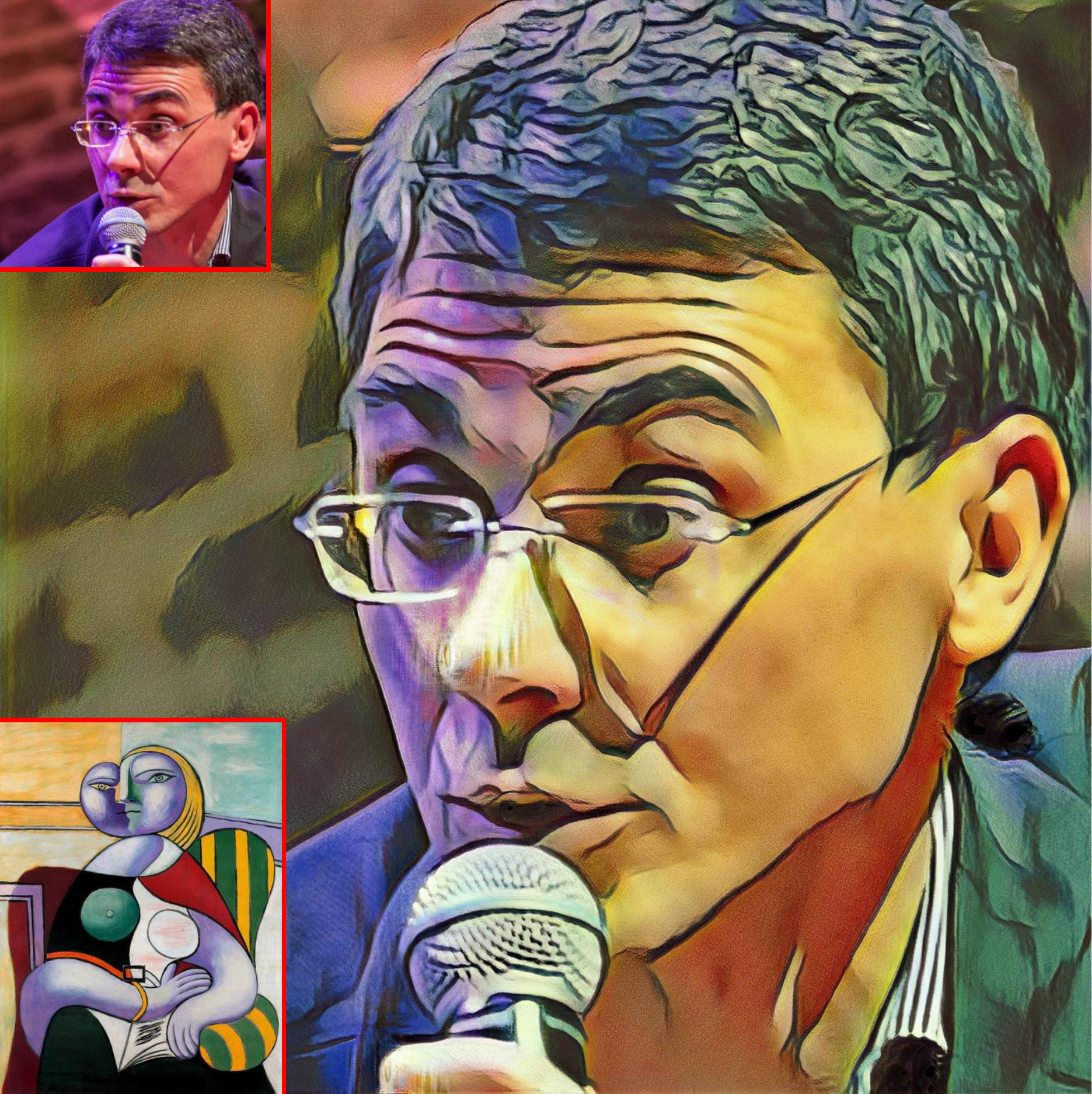}
\end{center}
\vspace{-3 mm}
  \caption{Picasso}
\label{Portrain_picasso}
\end{figure*}

\begin{figure*}
\begin{center}
\includegraphics[width=1.0\linewidth]{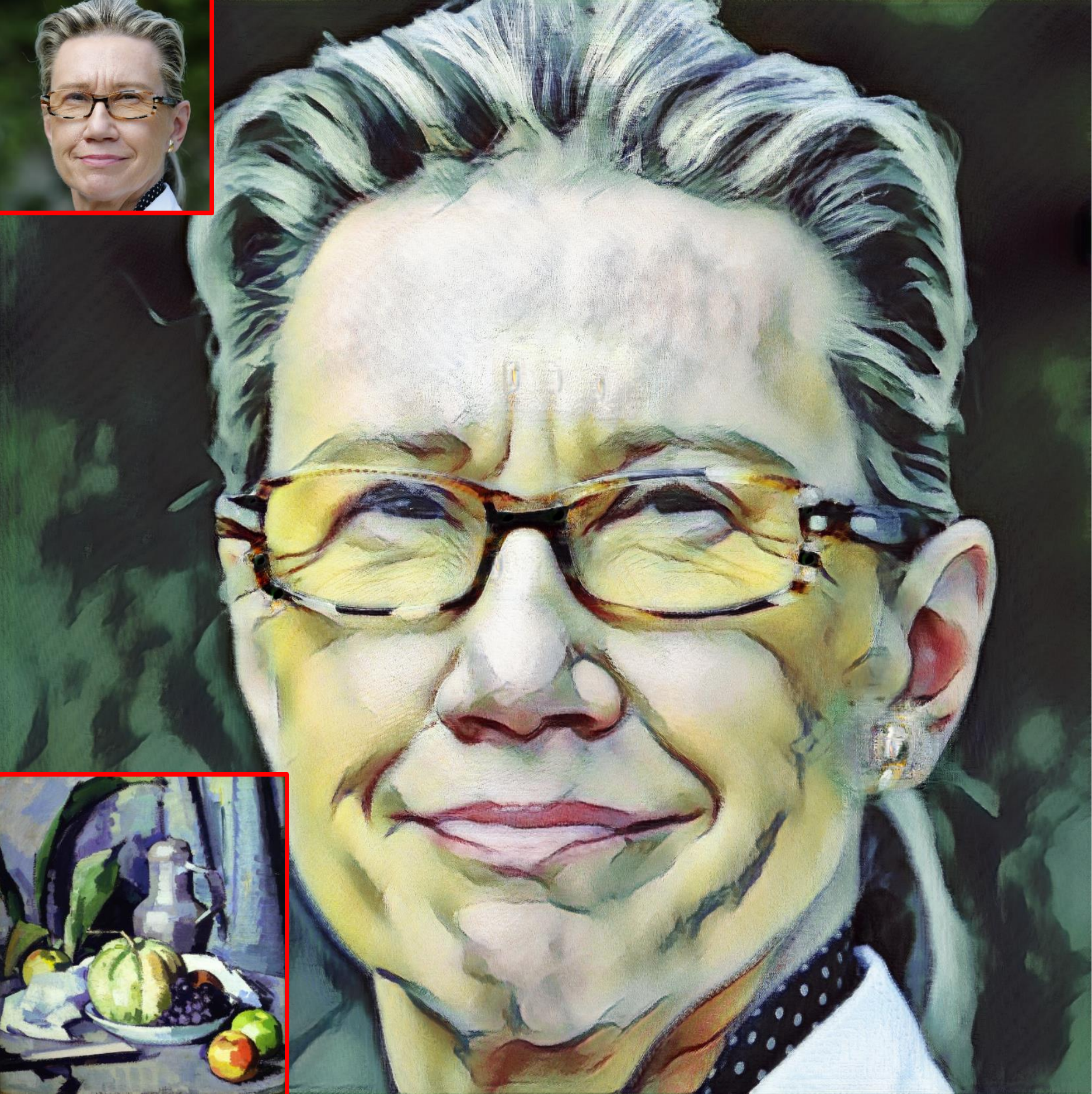}
\end{center}
\vspace{-3 mm}
  \caption{Samuel}
\label{Portrain_samuel1}
\end{figure*}

\begin{figure*}
\begin{center}
\includegraphics[width=1.0\linewidth]{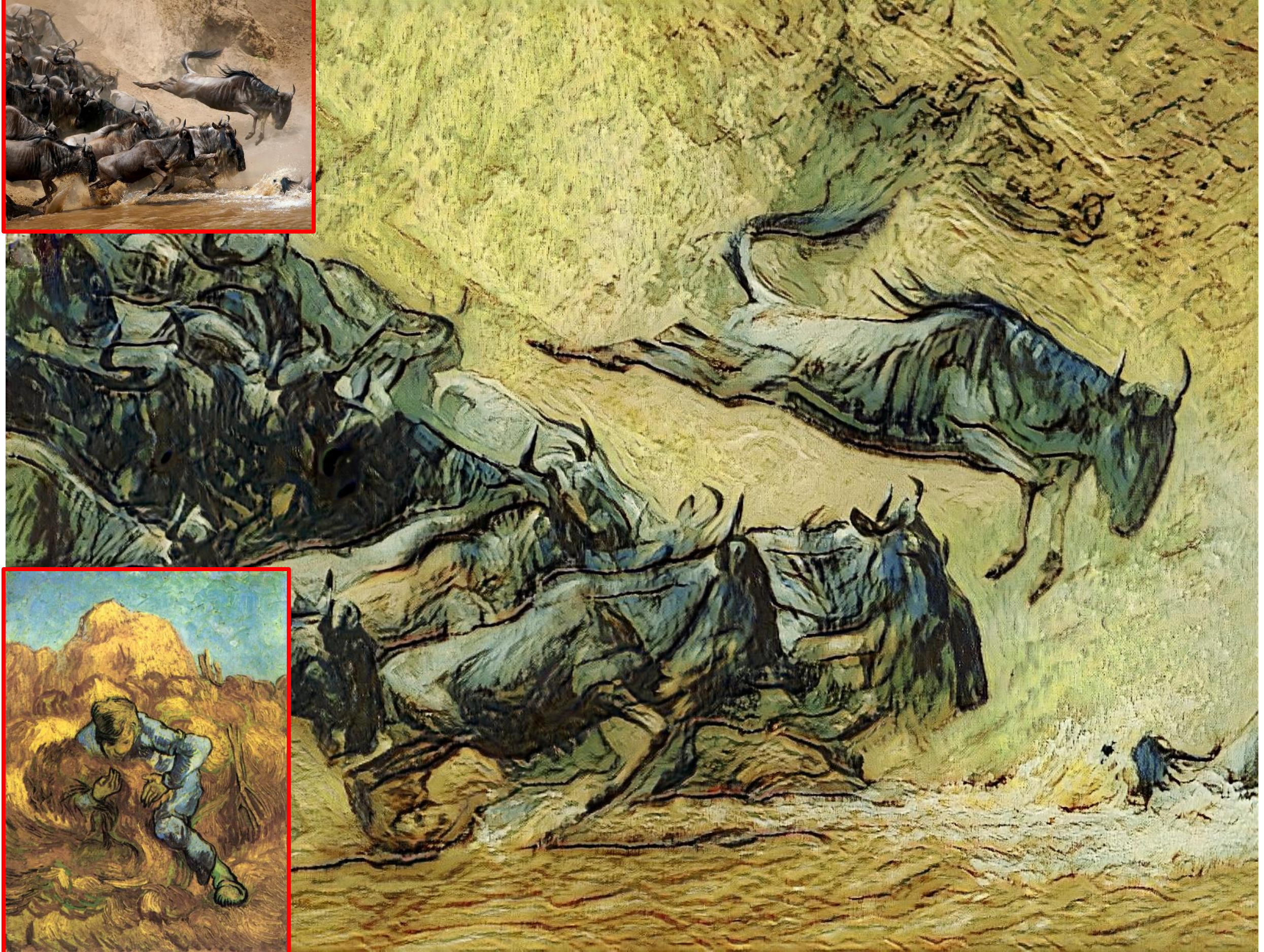}
\end{center}
\vspace{-3 mm}
  \caption{Vangogh}
\label{results2_vangogh2}
\end{figure*}

We provide more generated HD paintings using our method under eleven artists styles. Each generated image has a minimal side size of 1600 pixels. In Fig.~\ref{Portrain_picasso} to Fig.~\ref{results2_vangogh2}, we show the results of style transfer of eleven painter styles using ASMA-GAN. We select multiple types of pictures including portraits, woods, mountains, and buildings to show the generalization and flexibility of our method.

For each picture, the name of the picture is corresponding to the artist, the small picture in the upper left corner is the original picture, and the small picture in the lower left corner is one of the representative works of the artist.

From the results, we can see that the proposed ASMA-GAN realizes the style transfer while retaining the anisotropic semantic information. On one hand, the generated images realistically restore the original painter's style. By enlarging the picture, we can observe that the proposed method not only changes the color of the picture, but also imitates the stroke of the artists, which is the really meaning of style. And the differences between the generated styles of different artists are big by using just single network we proposed. On the other hand, the semantic information of the original image is easily discarded because of the excessive transfer of style. Our proposed method achieves a compromise between style and content. The content retained in the generated picture is consistent with the artist's drawing habits.

\begin{figure*}
\begin{center}
\includegraphics[width=1.0\linewidth]{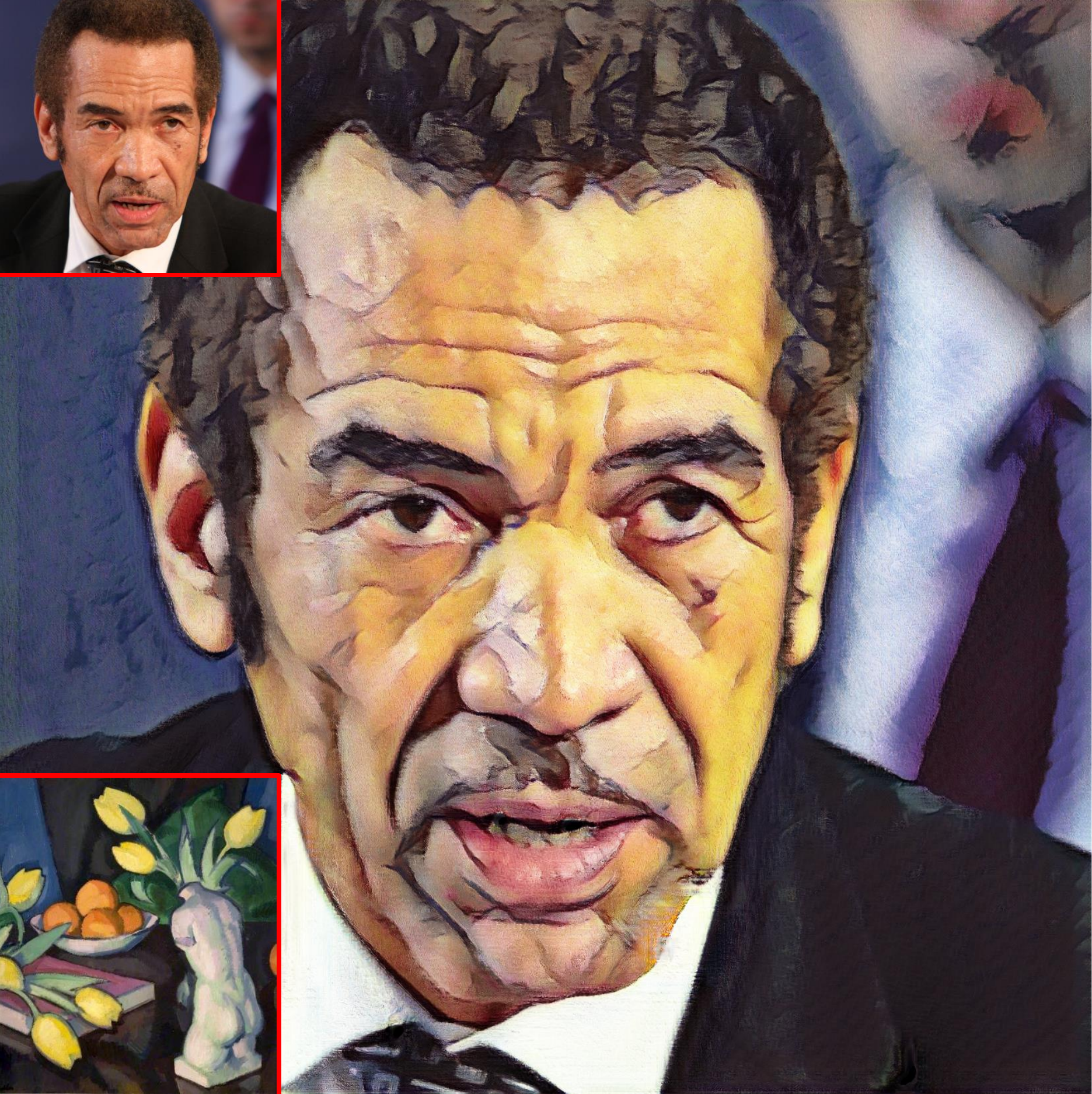}
\end{center}
\vspace{-3 mm}
  \caption{Samuel}
\label{Picasso}
\end{figure*}

\begin{figure*}
\begin{center}
\includegraphics[width=1.0\linewidth]{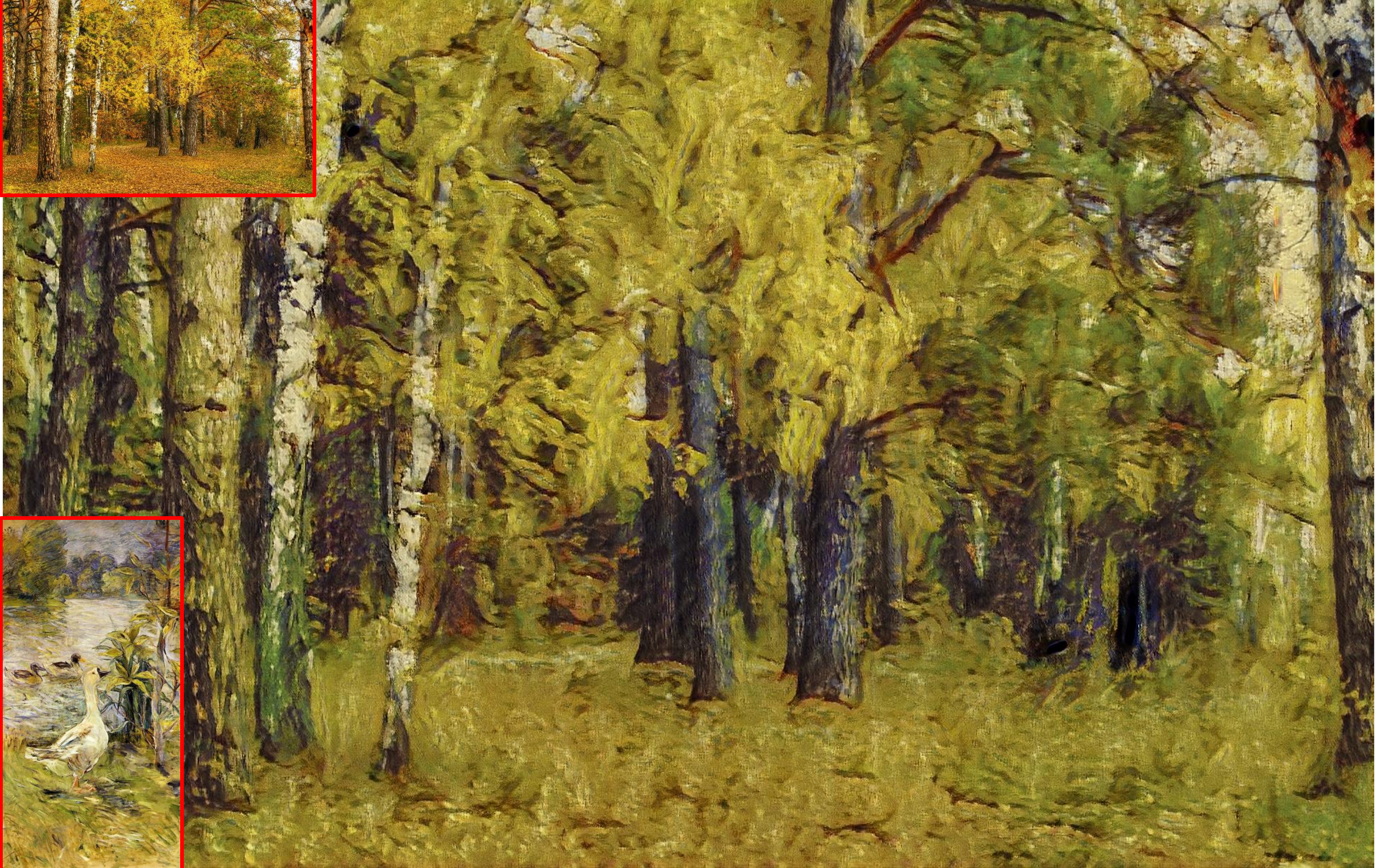}
\end{center}
\vspace{-3 mm}
  \caption{Morisot}
\label{Picasso}
\end{figure*}

\begin{figure*}
\begin{center}
\includegraphics[width=1.0\linewidth]{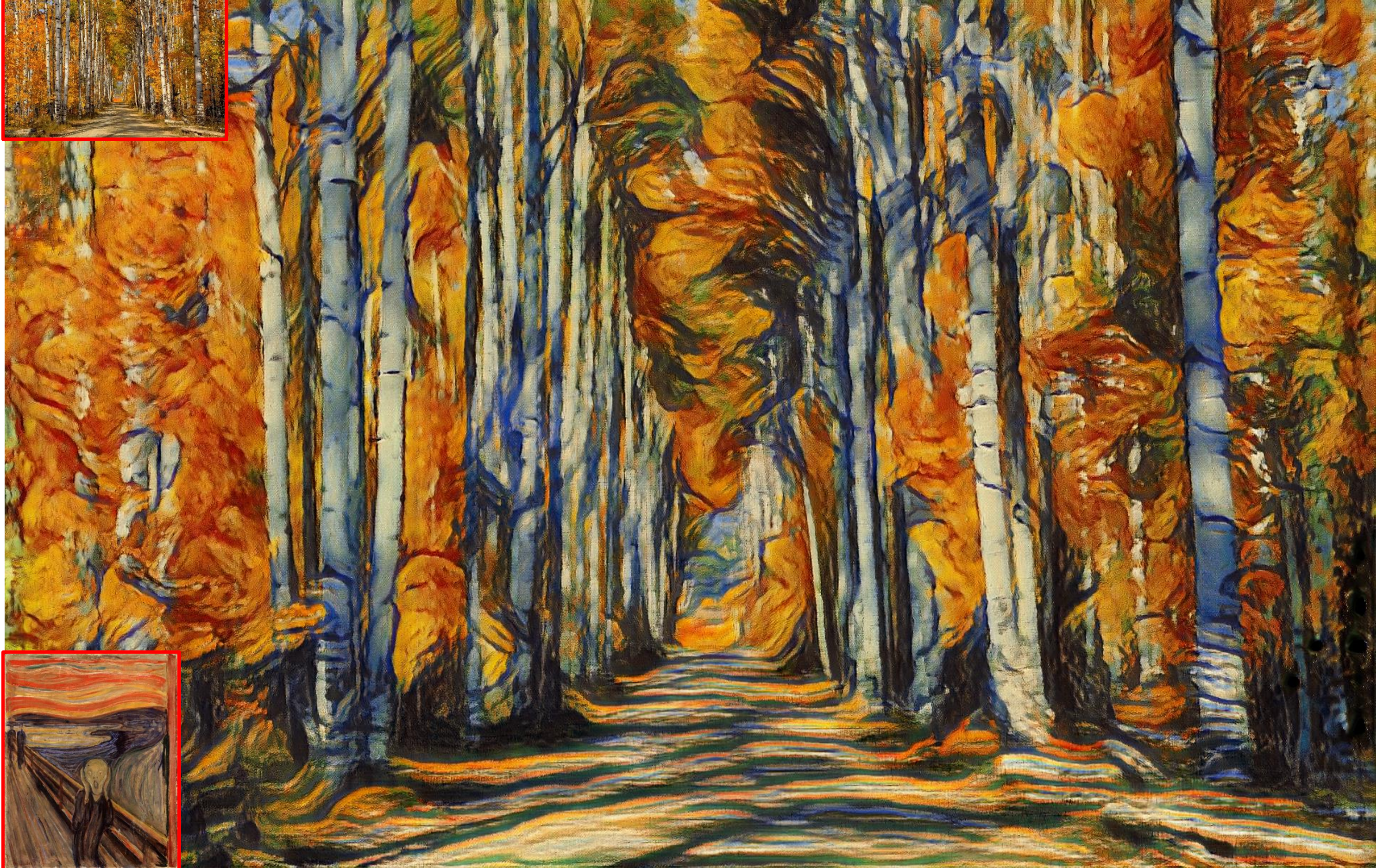}
\end{center}
\vspace{-3 mm}
  \caption{Munch}
\label{Picasso}
\end{figure*}

\begin{figure*}
\begin{center}
\includegraphics[width=1.0\linewidth]{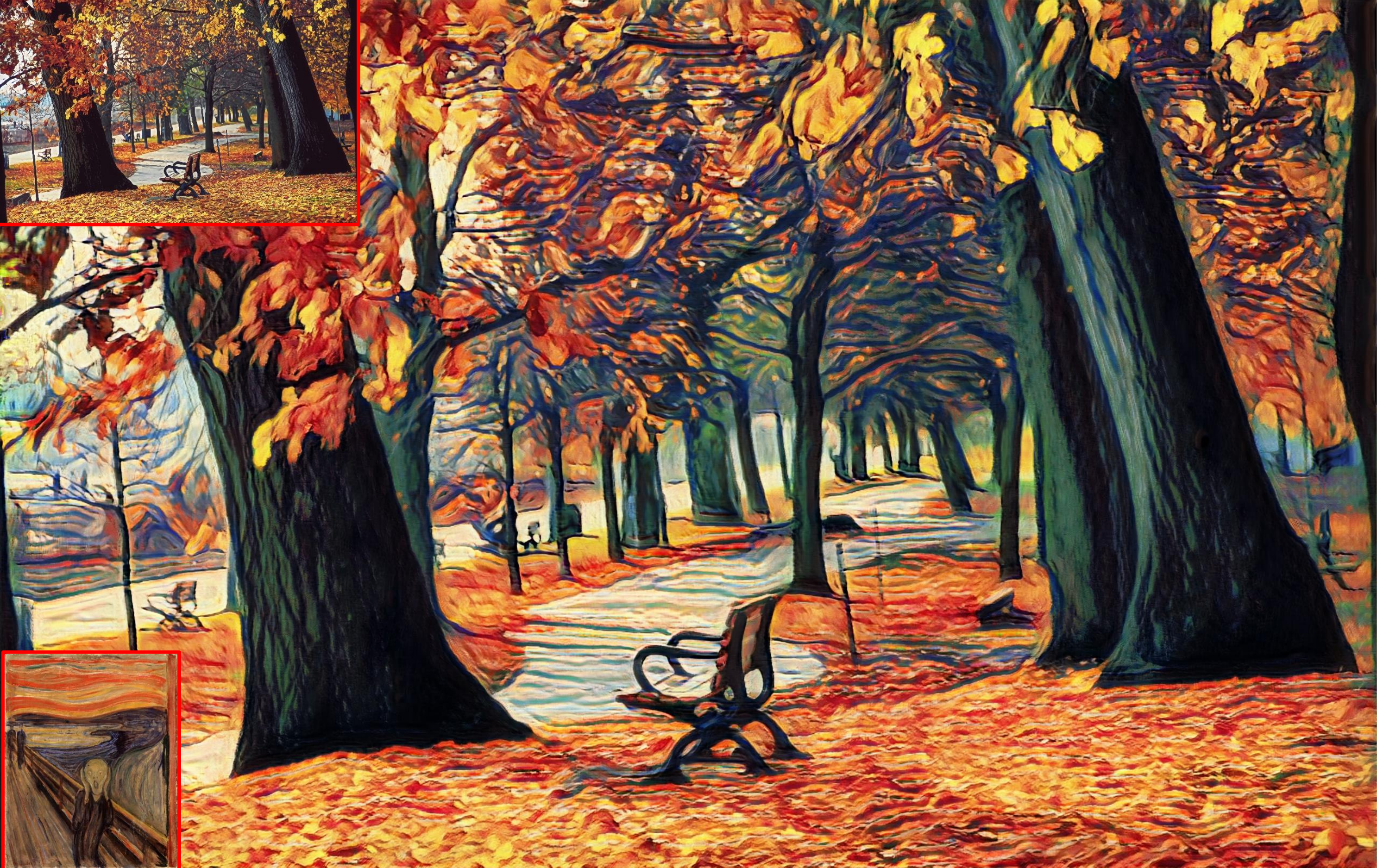}
\end{center}
\vspace{-3 mm}
  \caption{Munch}
\label{Picasso}
\end{figure*}

\begin{figure*}
\begin{center}
\includegraphics[width=1.0\linewidth]{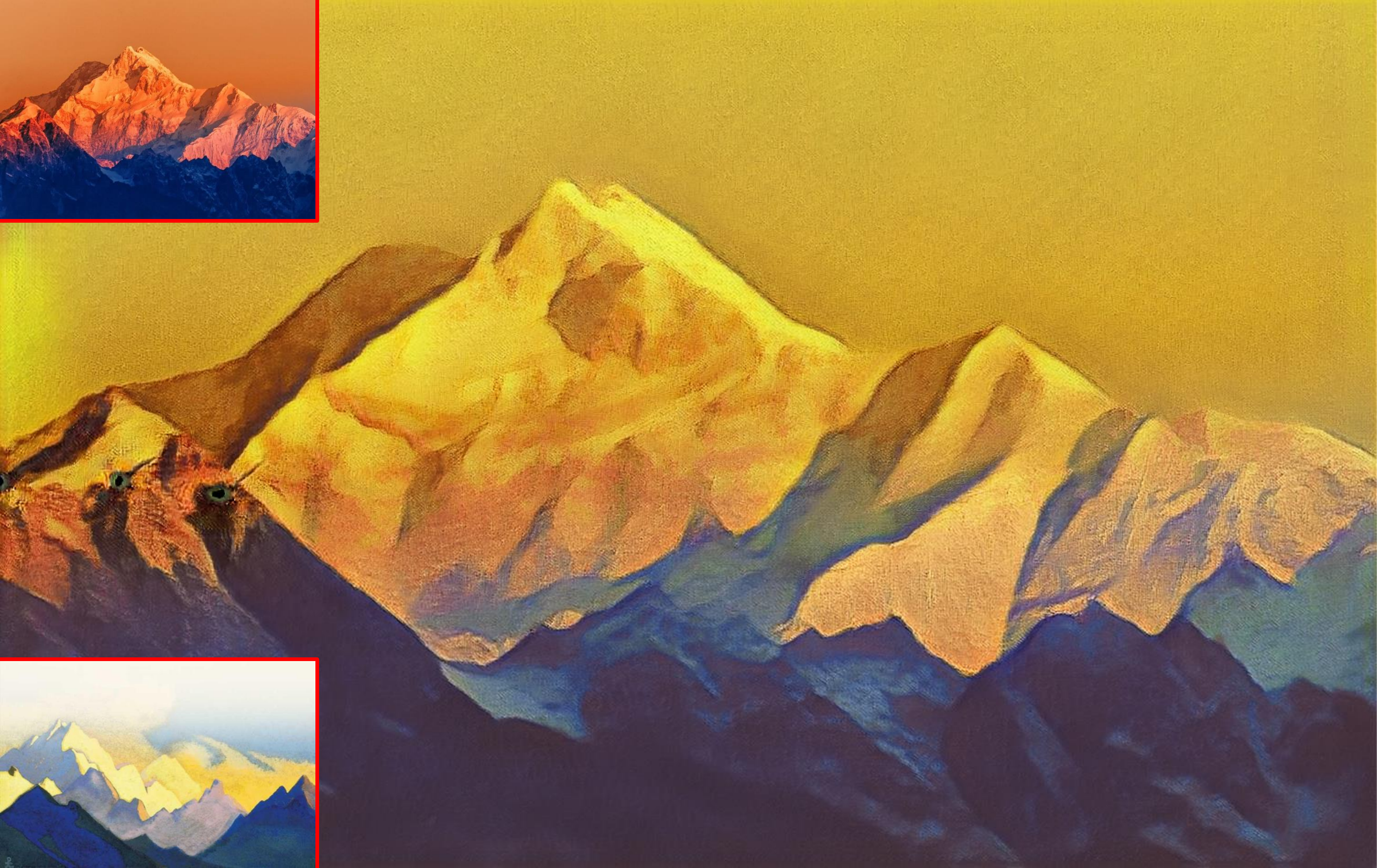}
\end{center}
\vspace{-3 mm}
  \caption{Nicholas}
\label{Picasso}
\end{figure*}

\begin{figure*}
\begin{center}
\includegraphics[width=1.0\linewidth]{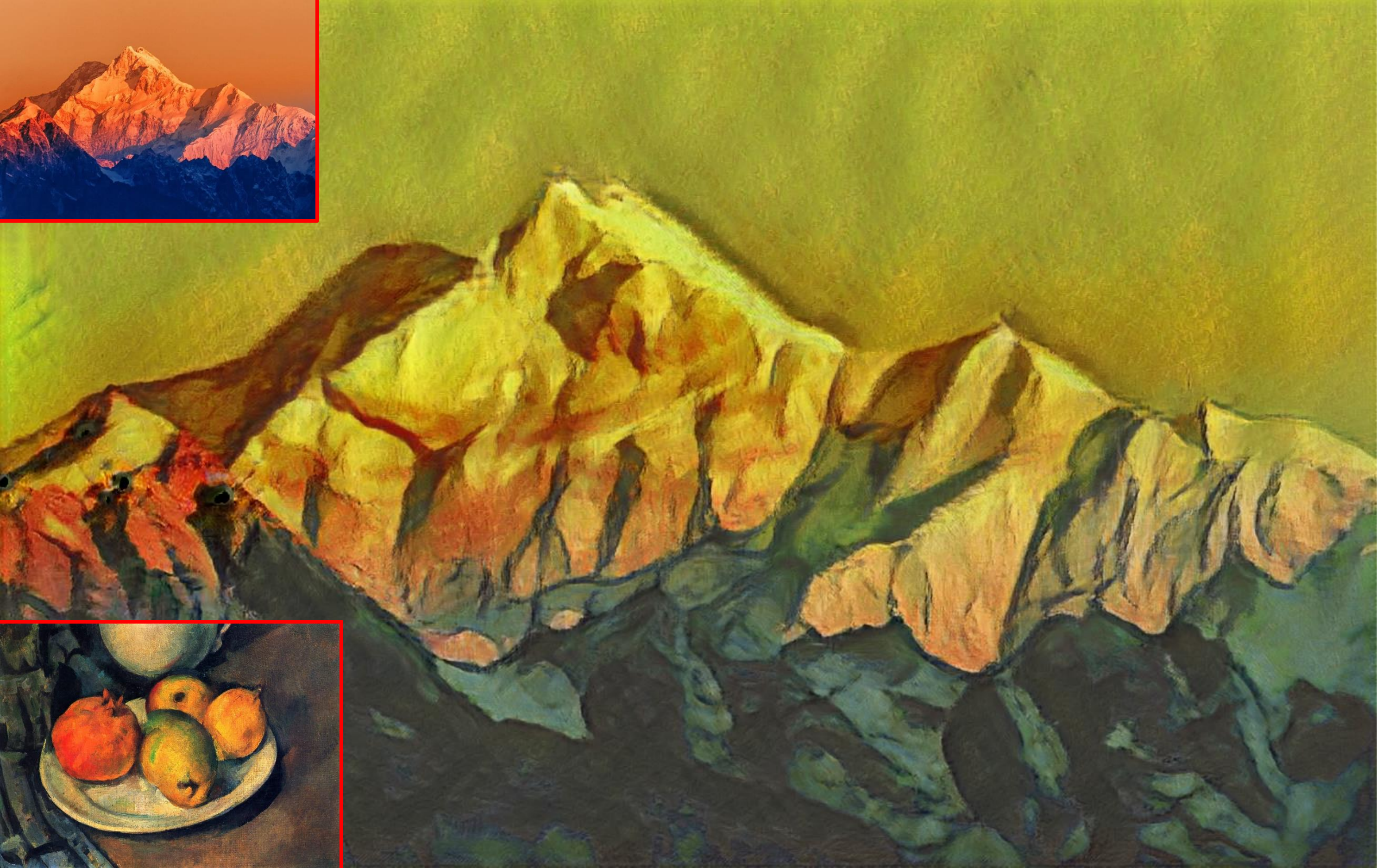}
\end{center}
\vspace{-3 mm}
  \caption{Cezanne}
\label{Picasso}
\end{figure*}

\begin{figure*}
\begin{center}
\includegraphics[width=1.0\linewidth]{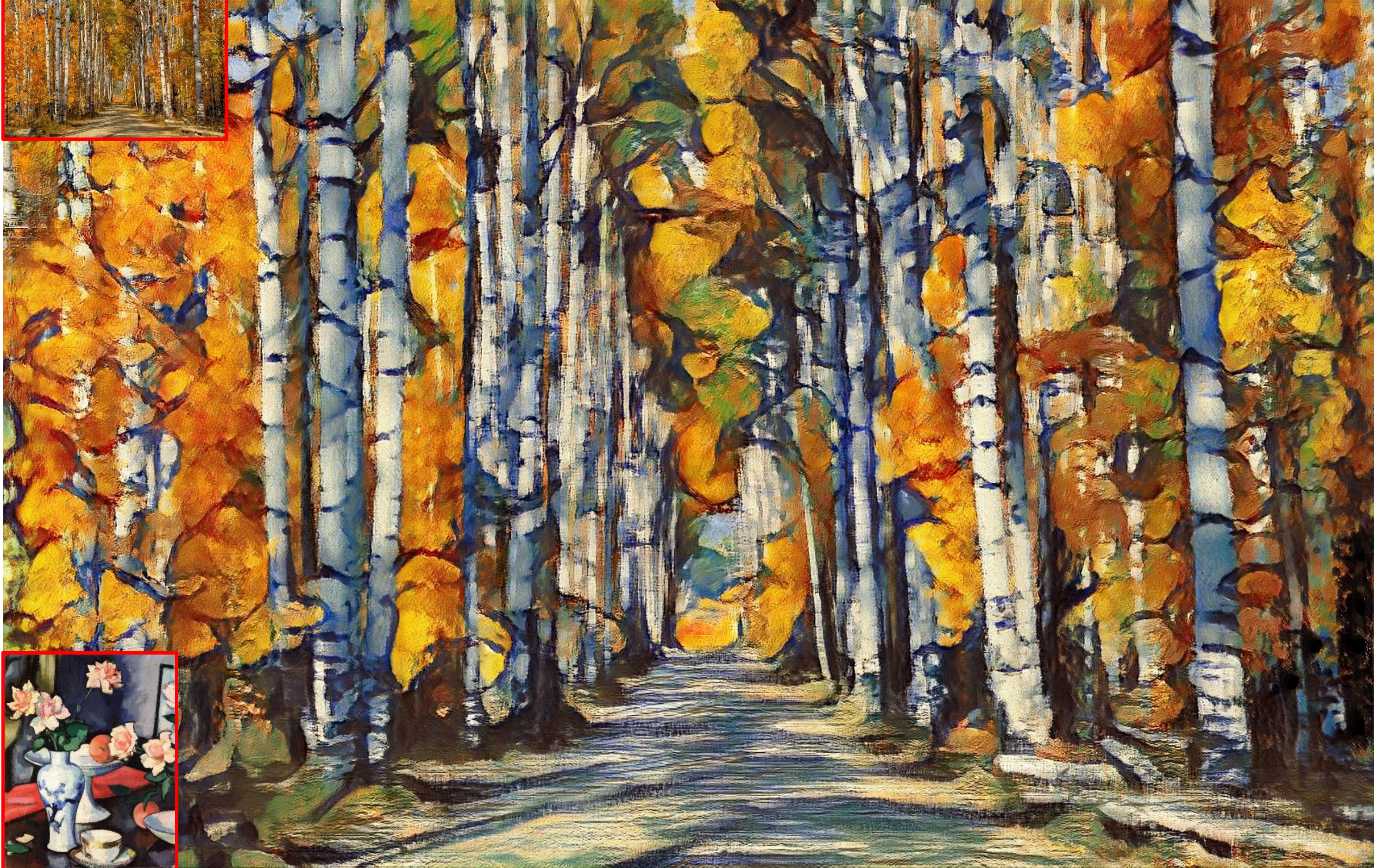}
\end{center}
\vspace{-3 mm}
  \caption{Samuel}
\label{Picasso}
\end{figure*}

\begin{figure*}
\begin{center}
\includegraphics[width=1.0\linewidth]{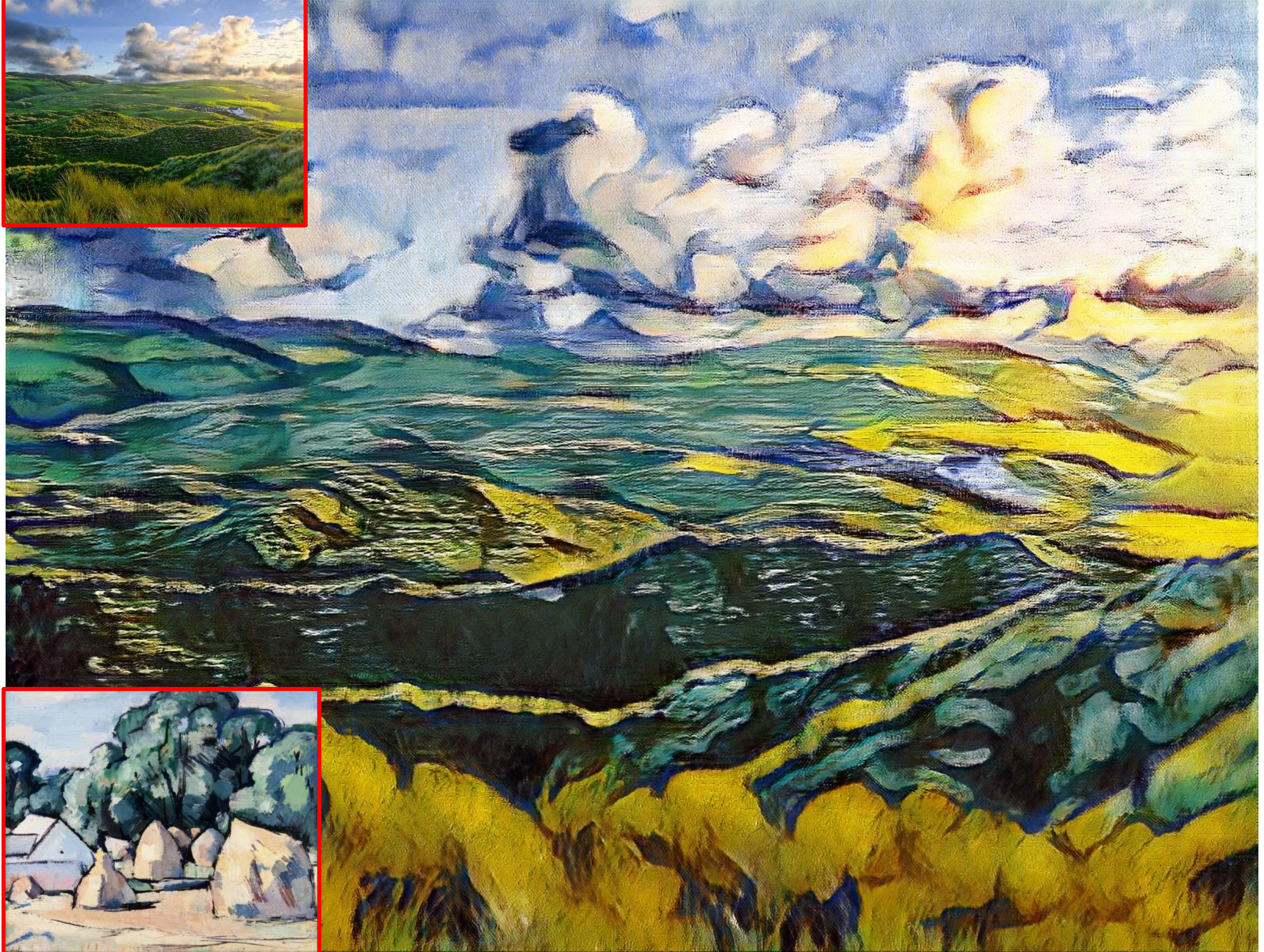}
\end{center}
\vspace{-3 mm}
  \caption{Samuel}
\label{Picasso}
\end{figure*}

\begin{figure*}
\begin{center}
\includegraphics[width=1.0\linewidth]{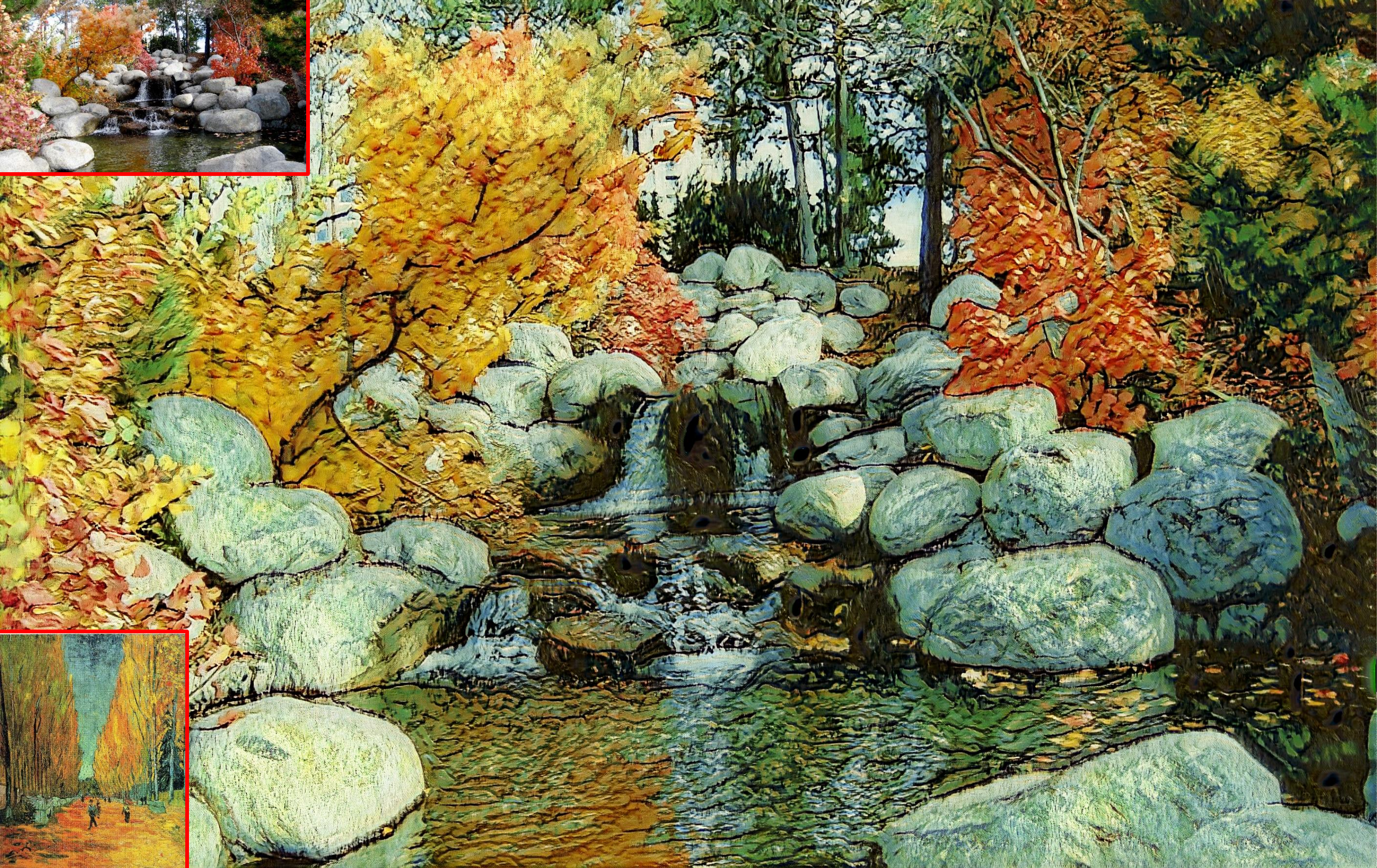}
\end{center}
\vspace{-3 mm}
  \caption{Vangogh}
\label{Picasso}
\end{figure*}

\begin{figure*}
\begin{center}
\includegraphics[width=1.0\linewidth]{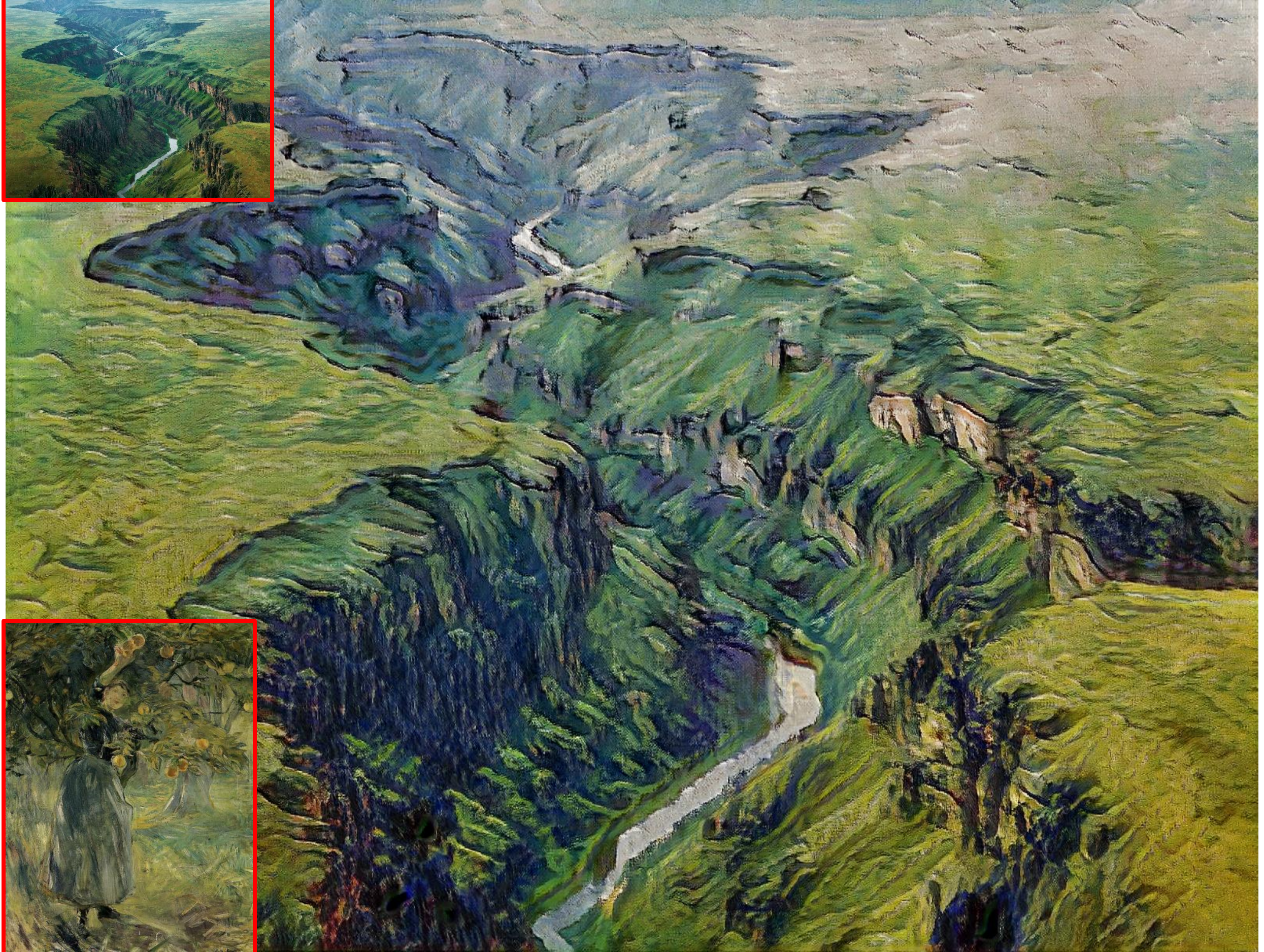}
\end{center}
\vspace{-3 mm}
  \caption{Morisot}
\label{Picasso}
\end{figure*}

\begin{figure*}
\begin{center}
\includegraphics[width=1.0\linewidth]{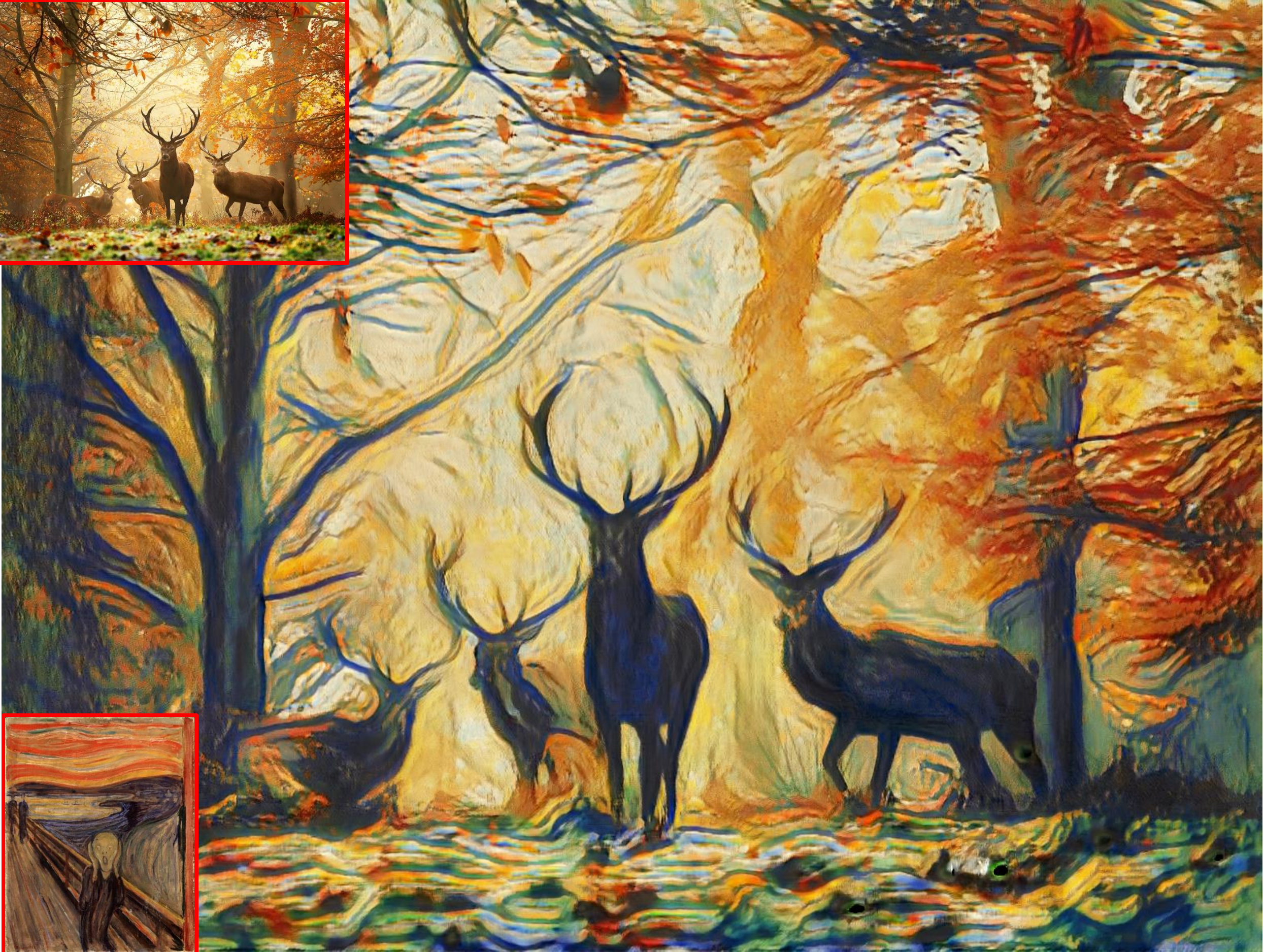}
\end{center}
\vspace{-3 mm}
  \caption{Munch}
\label{Picasso}
\end{figure*}

\begin{figure*}
\begin{center}
\includegraphics[width=1.0\linewidth]{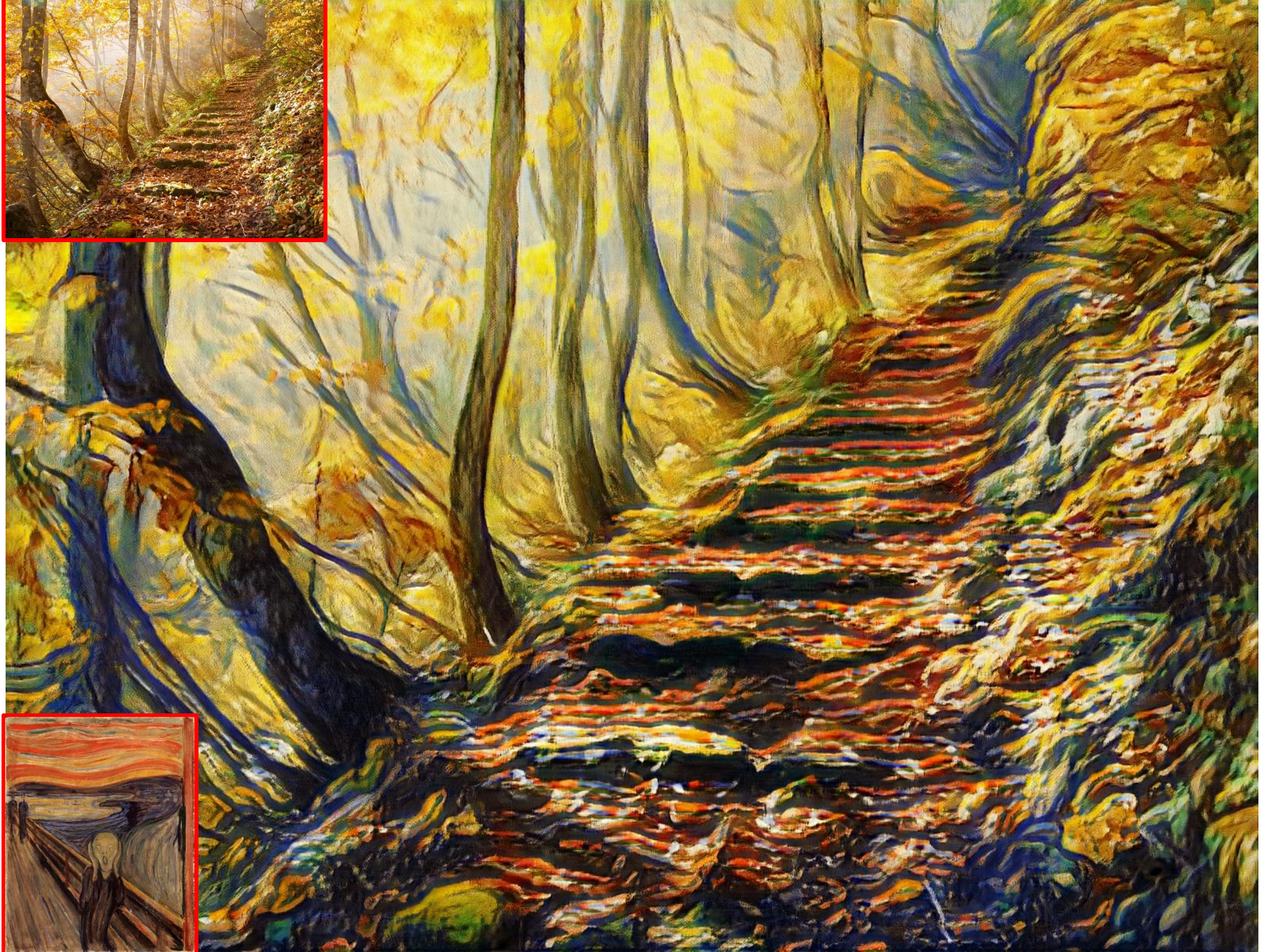}
\end{center}
\vspace{-3 mm}
  \caption{Munch}
\label{Picasso}
\end{figure*}

\begin{figure*}
\begin{center}
\includegraphics[width=1.0\linewidth]{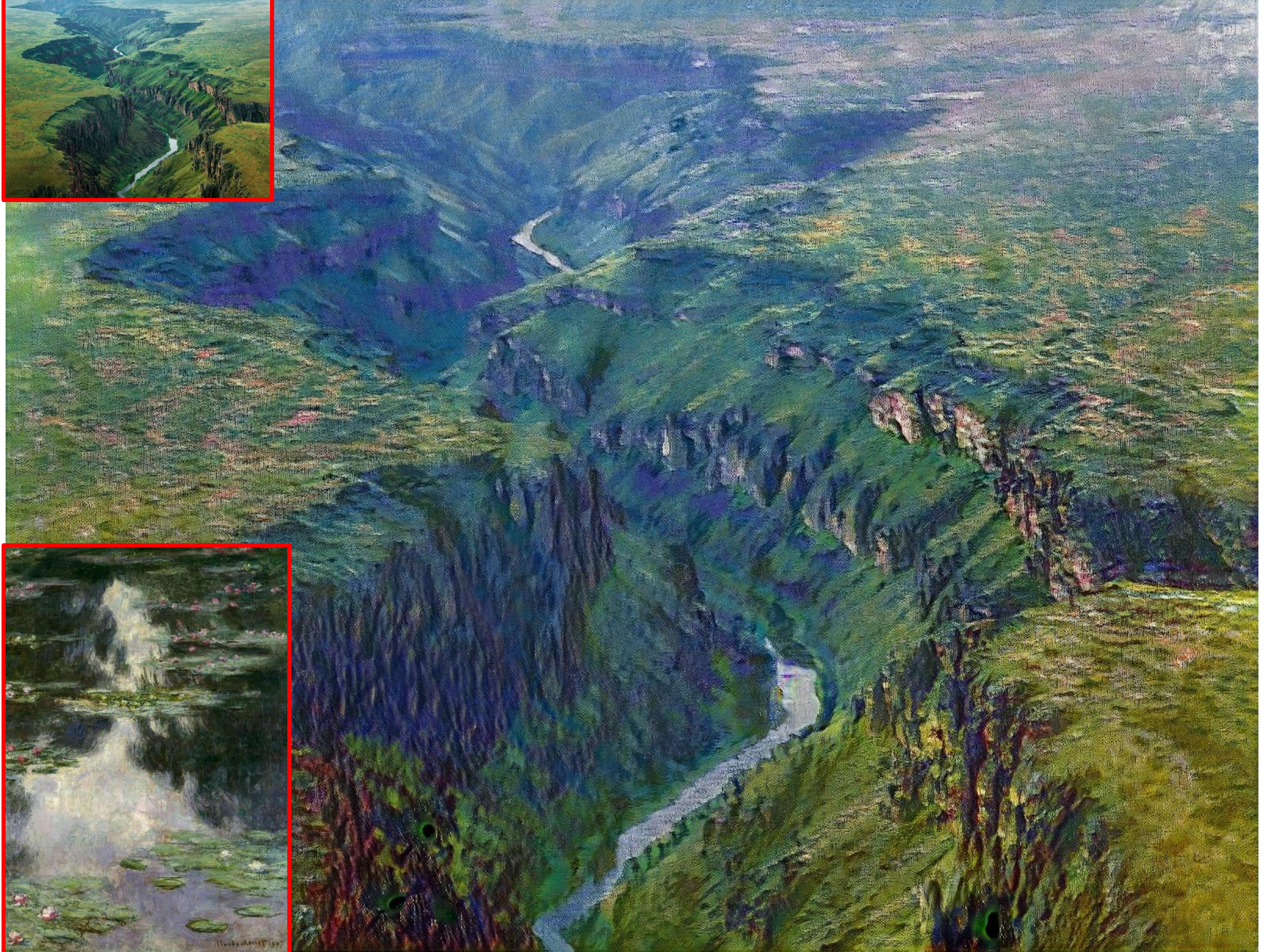}
\end{center}
\vspace{-3 mm}
  \caption{Monet}
\label{Picasso}
\end{figure*}

\begin{figure*}
\begin{center}
\includegraphics[width=1.0\linewidth]{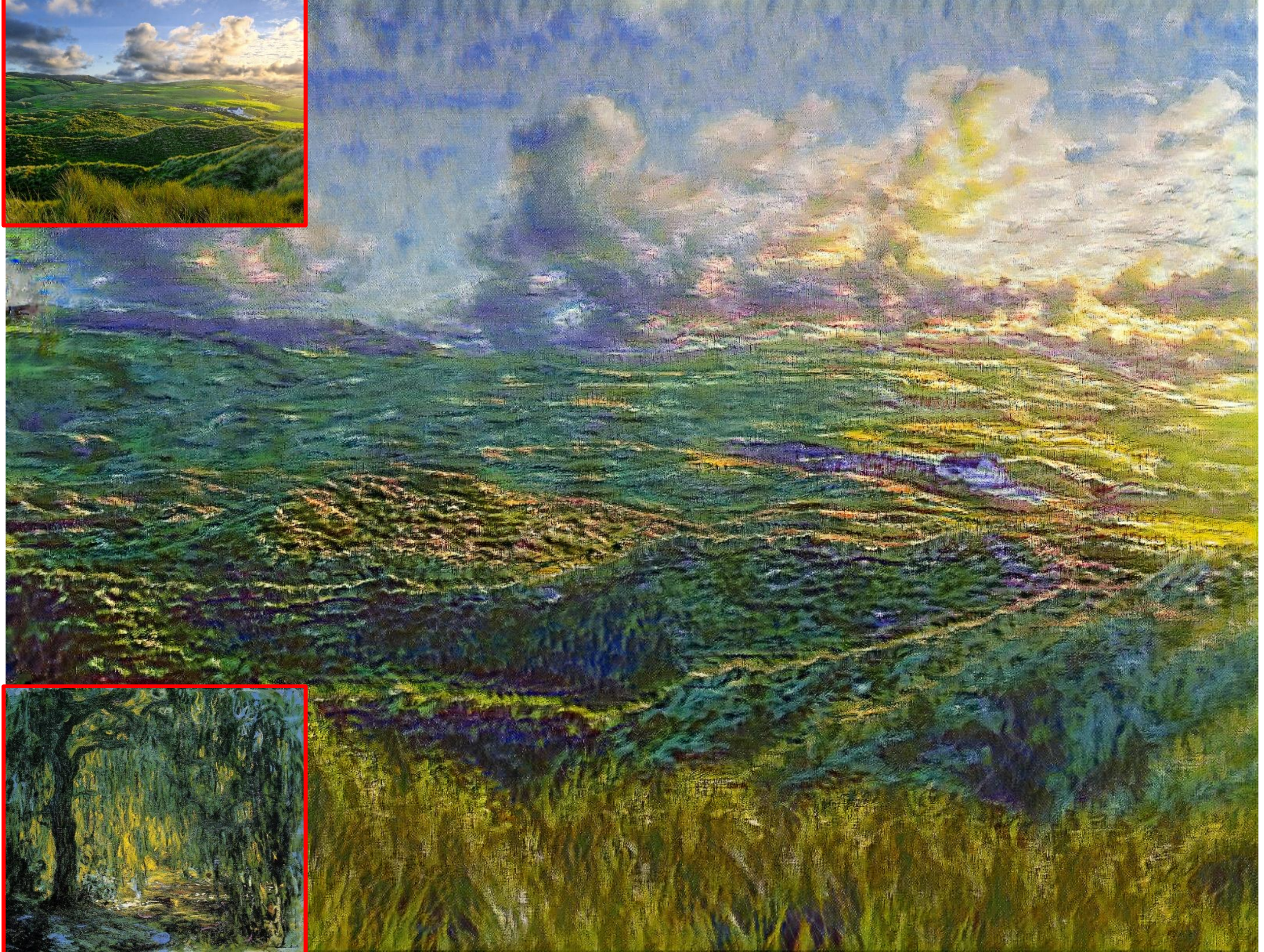}
\end{center}
\vspace{-3 mm}
  \caption{Monet}
\label{Picasso}
\end{figure*}

\begin{figure*}
\begin{center}
\includegraphics[width=1.0\linewidth]{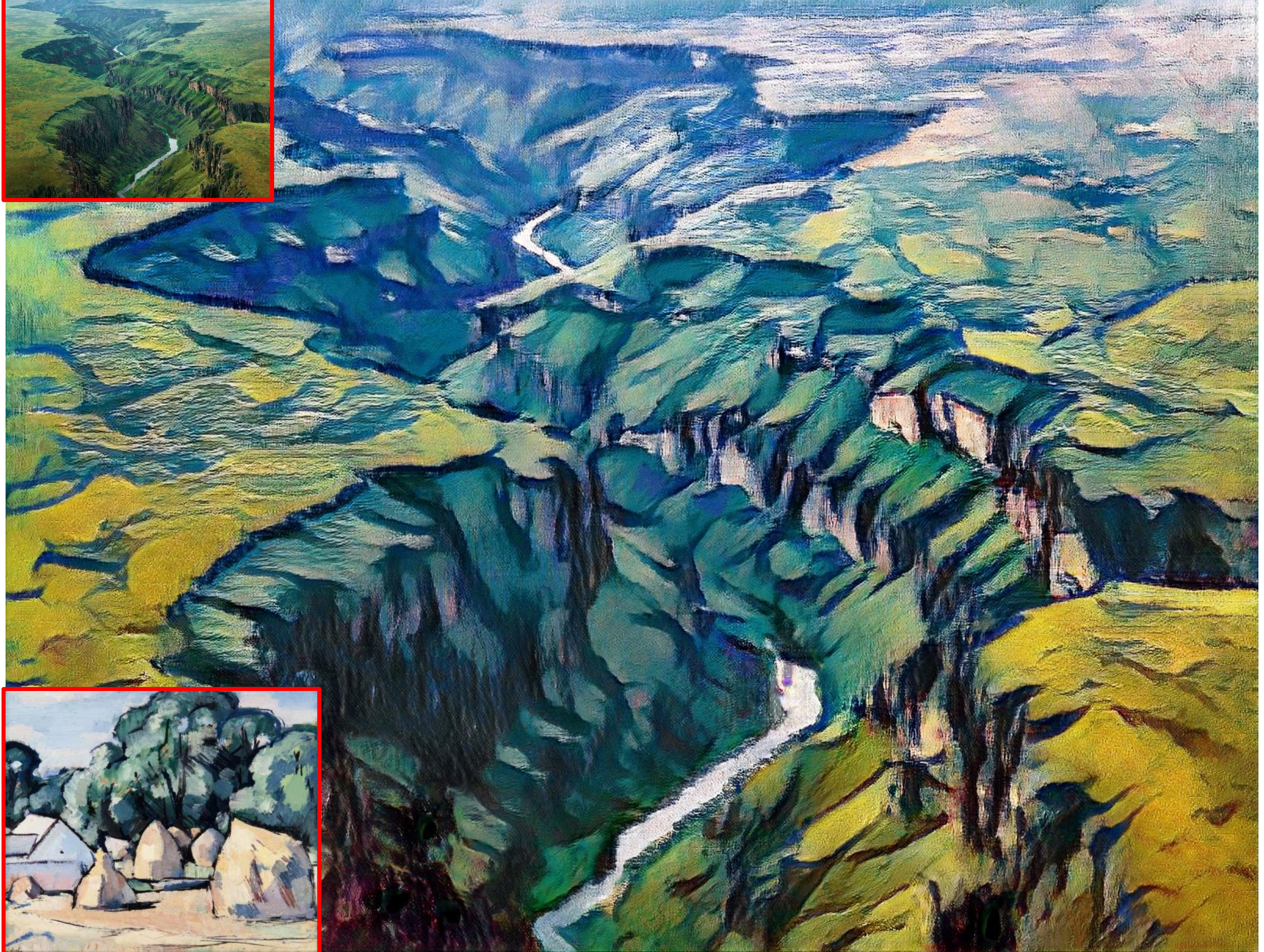}
\end{center}
\vspace{-3 mm}
  \caption{Samuel}
\label{Picasso}
\end{figure*}

\begin{figure*}
\begin{center}
\includegraphics[width=1.0\linewidth]{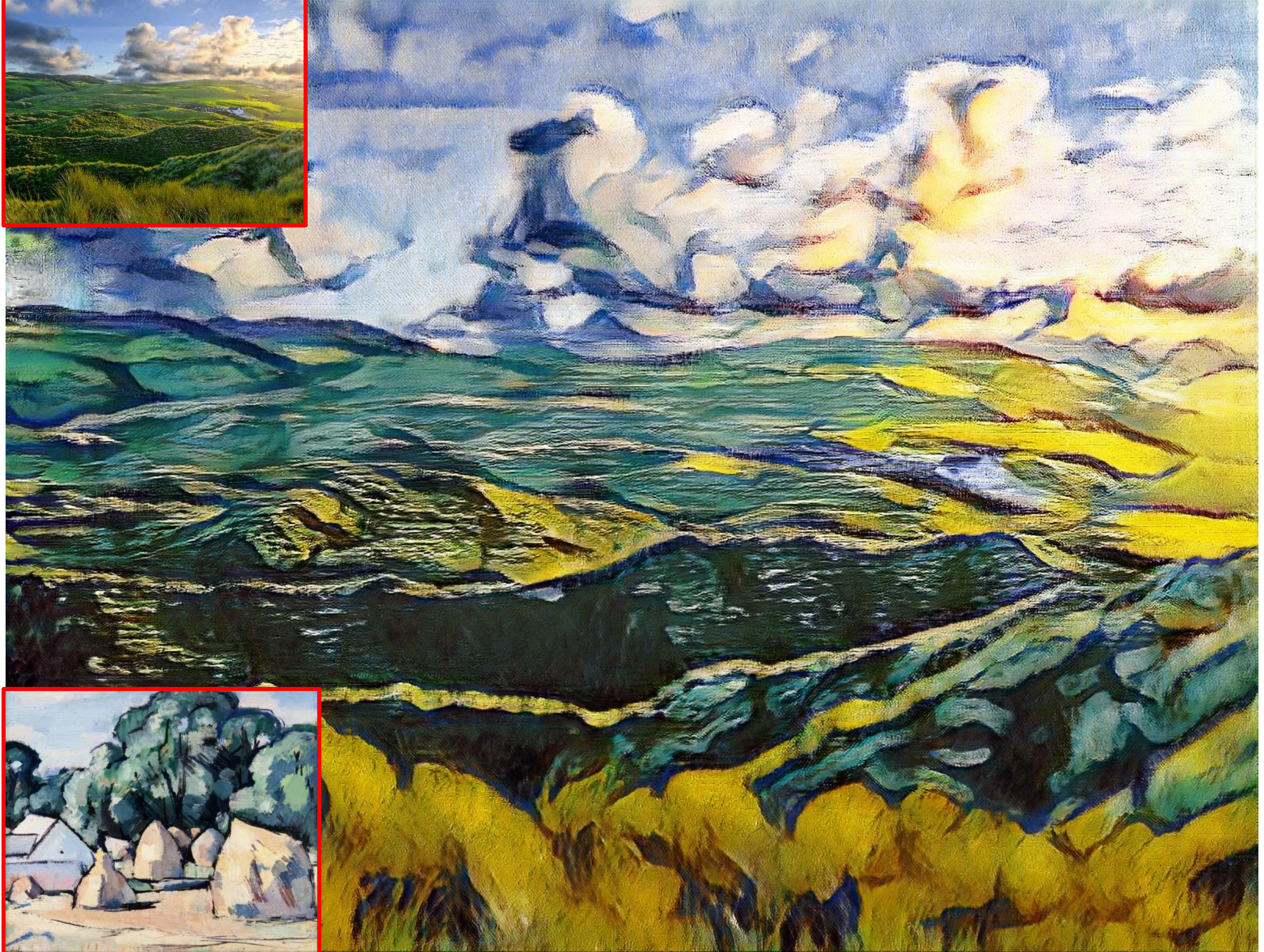}
\end{center}
\vspace{-3 mm}
  \caption{Samuel}
\label{Picasso}
\end{figure*}

\begin{figure*}
\begin{center}
\includegraphics[width=1.0\linewidth]{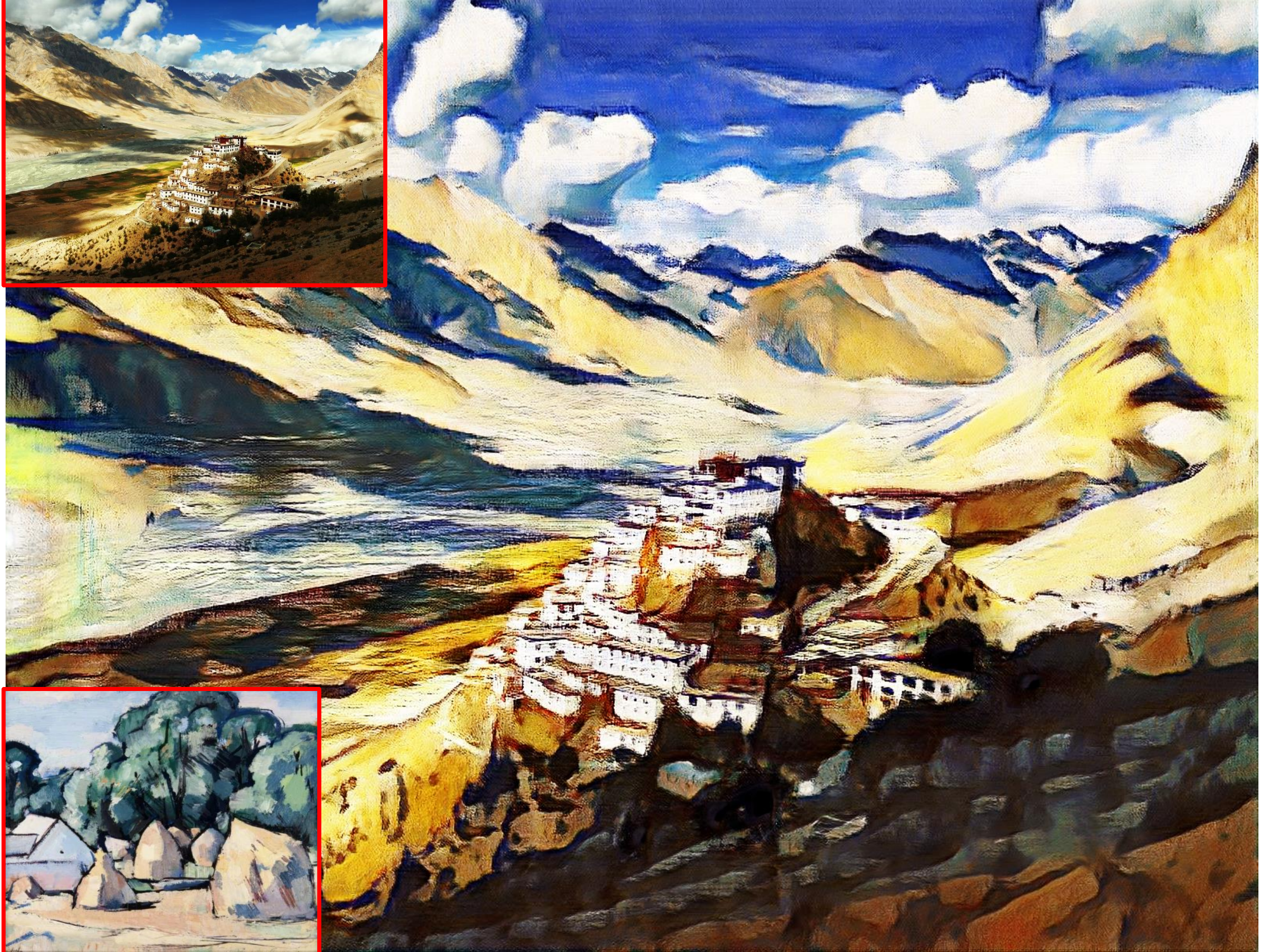}
\end{center}
\vspace{-3 mm}
  \caption{Samuel}
\label{Picasso}
\end{figure*}

\begin{figure*}
\begin{center}
\includegraphics[width=1.0\linewidth]{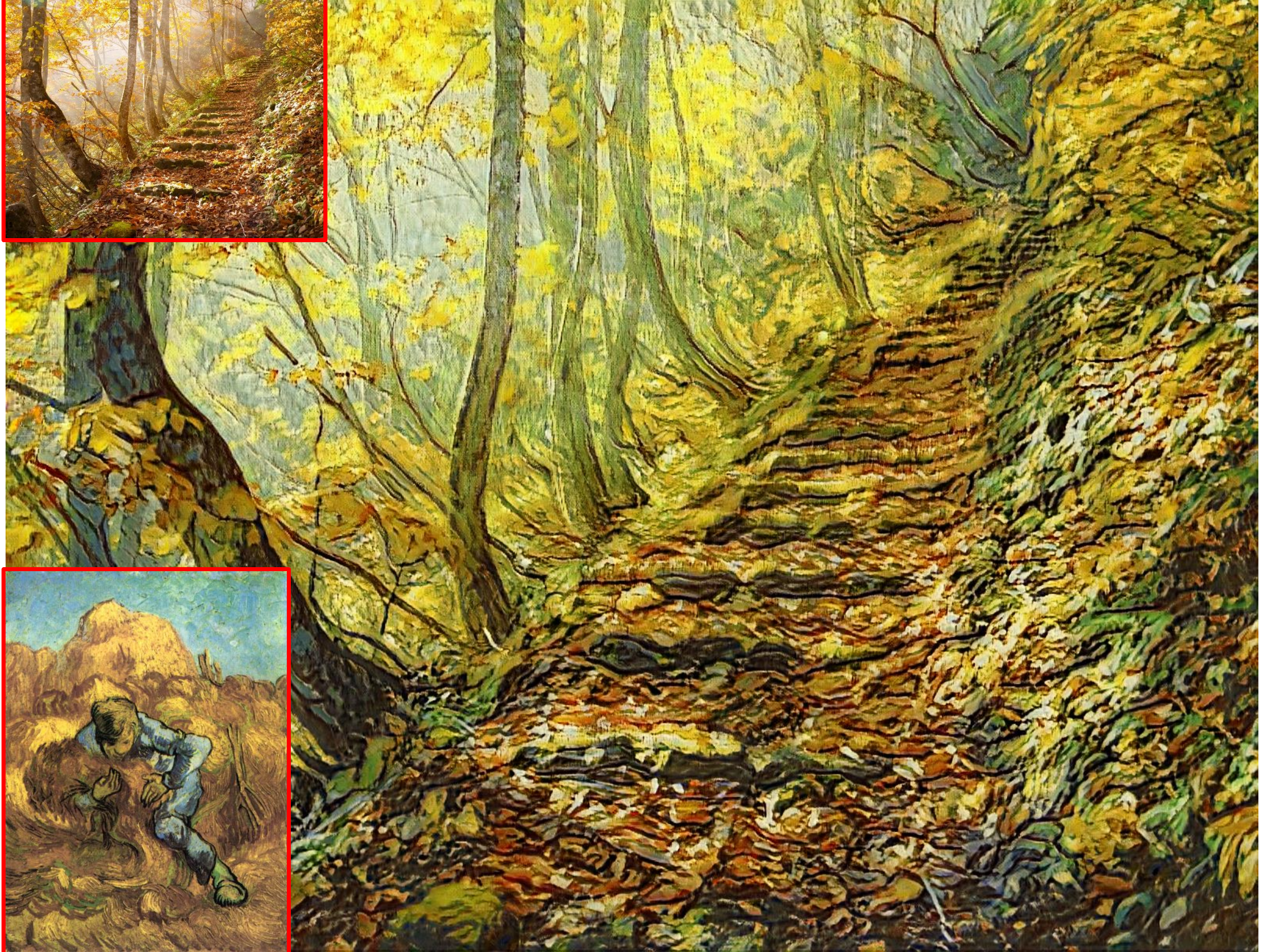}
\end{center}
\vspace{-3 mm}
  \caption{vangogh}
\label{Picasso}
\end{figure*}

\end{document}